\documentclass[11pt]{article}

\usepackage[final]{acl}

\usepackage{times}
\usepackage{latexsym}

\usepackage[T1]{fontenc}

\usepackage[utf8]{inputenc}

\usepackage{microtype}

\usepackage{inconsolata}

\usepackage{graphicx}

\usepackage{xspace}
\def\method{{\fontfamily{lmtt}\selectfont {\textbf{SwiAttn}}}\xspace}
\usepackage{pifont}
\usepackage{enumitem}
\usepackage{bm}
\usepackage{amssymb,amsfonts}
\usepackage{mathtools}
\DeclarePairedDelimiter{\norm}{\lVert}{\rVert}
\usepackage{dsfont}
\usepackage[most]{tcolorbox}
\usepackage{xcolor}
\usepackage{colortbl}
\usepackage{multirow}
\usepackage{makecell}
\usepackage{hhline}
\usepackage{float}
\usepackage{placeins}
\usepackage{balance}
\usepackage{natbib}
\usepackage{algorithm}
\usepackage{algorithmic}
\usepackage{caption}
\usepackage{ragged2e}
\usepackage{subcaption}
\usepackage{listings}
\definecolor{tableheadcolor}{RGB}{137, 206, 228}
\definecolor{tablerowcolor}{RGB}{242, 242, 242}
\definecolor{ourrowname}{RGB}{245, 231, 172}
\definecolor{codekeyword}{RGB}{112, 63, 147}
\definecolor{codecomment}{RGB}{55, 124, 90}
\definecolor{codestring}{RGB}{159, 73, 66}
\definecolor{codenumber}{RGB}{105, 105, 105}
\lstdefinestyle{appendixcode}{
    language=Python,
    basicstyle=\ttfamily\scriptsize\linespread{0.90}\selectfont,
    keywordstyle=\color{codekeyword}\bfseries,
    commentstyle=\color{codecomment},
    stringstyle=\color{codestring},
    numberstyle=\tiny\color{codenumber},
    numbers=left,
    numbersep=6pt,
    frame=lines,
    framesep=1.5mm,
    breaklines=true,
    breakatwhitespace=false,
    columns=fullflexible,
    keepspaces=true,
    showstringspaces=false,
    xleftmargin=1.5em,
    aboveskip=0pt,
    belowskip=0pt
}
\newcommand{\rev}[1]{\textcolor{black}{#1}}
\newcommand{\camrev}[1]{#1}
\newenvironment{camrevblock}{}{}

\title{Switch Attention: Towards Dynamic and Fine-grained Hybrid Transformers}

\author{
Yusheng Zhao\textsuperscript{$\heartsuit$ $\diamondsuit$},
Hourun Li\textsuperscript{$\heartsuit$ $\diamondsuit$},
Bohan Wu\textsuperscript{$\heartsuit$ $\diamondsuit$},
\textbf{Yichun Yin}\textsuperscript{\ding{168}}, \\
\textbf{Lifeng Shang}\textsuperscript{\ding{168}},
\textbf{Jingyang Yuan}\textsuperscript{$\heartsuit$ \textdagger},
\textbf{Meng Zhang}\textsuperscript{\ding{168} \textdagger},
\textbf{Ming Zhang}\textsuperscript{$\heartsuit$ \textdagger} \\
\textsuperscript{$\heartsuit$} State Key Laboratory for Multimedia Information Processing, School of Computer Science, \\
PKU-Anker LLM Lab, Peking University, Beijing, China \quad
\textsuperscript{\ding{168}} Huawei Technologies Co., Ltd. \\
\textsuperscript{$\diamondsuit$} Contribution during internship at Huawei Technologies Co., Ltd. \\
\texttt{\{yusheng.zhao,lihourun\}@stu.pku.edu.cn},
\texttt{\{wxtpku,yuanjy,mzhang\_cs\}@pku.edu.cn} \\
\texttt{\{zhangmeng92,yinyichun,Shang.Lifeng\}@huawei.com}
}

\makeatletter
\def\blfootnote{\xdef\@thefnmark{}\@footnotetext}
\makeatother

\begin{document}
\maketitle
\begin{abstract}
The attention mechanism has been the core component in modern transformer architectures. However, the computation of standard full attention scales quadratically with the sequence length, serving as a major bottleneck in long-context language modeling. Sliding window attention restricts the context length for better efficiency at the cost of narrower receptive fields. While existing efforts attempt to take the benefits from both sides by building hybrid models, they often resort to static, heuristically designed alternating patterns that limit efficient allocation of computation in various scenarios. 
In this paper, we propose Switch Attention (\method{}), a novel hybrid transformer that enables dynamic and fine-grained routing between full attention and sliding window attention. For each token at each transformer layer, \method{} dynamically routes the computation to either a full-attention branch for global information aggregation or a sliding-window branch for efficient local pattern matching. An adaptive regularization objective is designed to encourage the model towards efficiency. Moreover, we adopt continual pretraining to optimize the model, transferring the full attention architecture to the hybrid one.
Extensive experiments are conducted on twenty-three benchmark datasets across both regular (4K) and long (32K) context lengths, demonstrating the effectiveness of the proposed method. Our kernel implementation is available at \url{https://github.com/YushengZhao/SwitchAttentionKernels}.
\end{abstract}
\blfootnote{
\textsuperscript{\textdagger} Corresponding authors.
}

\section{Introduction}
The transformer architectures \cite{vaswani2017attention, dai2019transformer, kitaev2020reformer} have been widely adopted in recent large language models (LLMs) \cite{team2023gemini, openai2023gpt4, singh2025openai}. Despite their success in language modeling, standard transformer architectures suffer from quadratic computational complexity due to the attention mechanism \cite{vaswani2017attention}. To address this limitation, efforts have been made in reducing the computation of standard full attention \cite{shazeer2019fast, dao2022flashattention, ainslie2023gqa, liu2024deepseek}, sparsifying full attention \cite{child2019generating, kitaev2020reformer, liu2022dynamic, zhang2023h2o, xiao2023efficient, song2024low, xu2025xattention, yuan2025native}, developing sub-quadratic attention mechanisms \cite{katharopoulos2020transformers, ma2021luna, han2023flatten, peng2023rwkv, yang2024gated, gu2024mamba, peng2025rwkv, guo2025log}, and designing hybrid transformer models \cite{beltagy2020longformer, zaheer2020big, Jiang2023Mistral7, fu2024mixture, du2025native}.

\begin{figure}
    \centering
    \includegraphics[width=\linewidth]{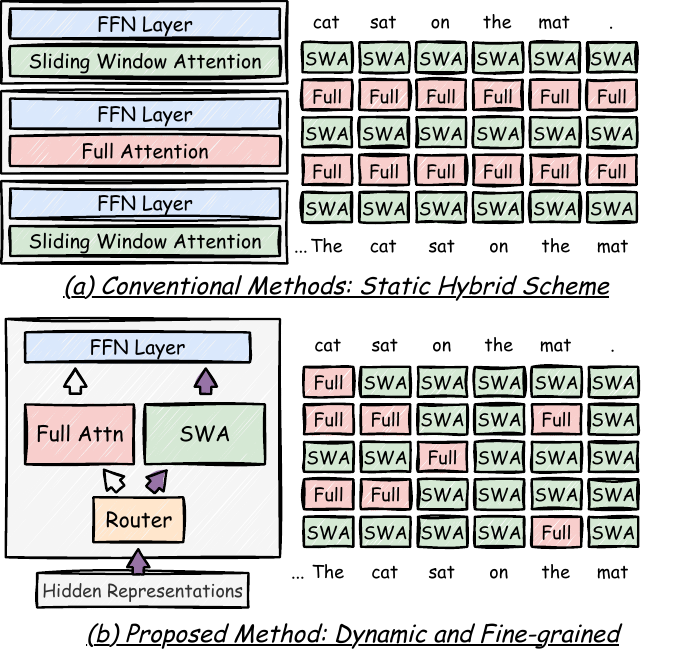}
    \vspace{-4mm}
    \caption{Conventional hybrid transformers (a) often adopt manually designed, static hybrid schemes. In contrast, the proposed \method{} (b) dynamically selects the type of attention for each input token at each transformer layer, enabling fine-grained allocation of computation.}
    \label{fig:motivation}
    \vspace{-4mm}
\end{figure}

Among these solutions, hybrid transformer models leverage both standard full attention and its efficient variants, \emph{e.g.}, sliding window attention (SWA, \citealt{beltagy2020longformer}), and have been widely adopted in larger industrial models \cite{team2024gemma, yang2025qwen3, openai2025gptoss, gao2026hysparse}. While they balance efficiency and performance of language modeling, most hybrid transformers adopt a \textbf{\textit{static hybrid scheme}} (\emph{e.g.}, alternating between full attention and sliding window attention with a fixed ratio) designed heuristically prior to the pretraining stage \cite{brown2020language, Jiang2023Mistral7, singh2025openai, du2025native}. This limits the model's potential to allocate computation more efficiently. For example, when the context beyond the range of a sliding window is less relevant to the response that the model is about to generate, one would expect a lower ratio of expensive standard full attention, and vice versa.

Towards this end, this paper proposes a novel hybrid attention mechanism named Switch Attention (\method{}). As is shown in Figure \ref{fig:motivation}, the proposed \method{} is a \textbf{\textit{dynamic and fine-grained}} hybrid solution compared to prior efforts. In particular, \method{} inserts a router before each attention module and dynamically decides whether the input feature should be processed by standard full attention or sliding window attention for each token at each transformer layer. The full attention branch gathers long-range information accurately from the entire sequence, while the sliding window branch models short-term memory efficiently from a narrower context. The router dynamically decides the specific branch for each token at each layer based on the input feature. In order to encourage the router towards the more efficient branch (\emph{i.e.}, the SWA branch), we adopt a regularization objective that is optimized with the log-likelihood loss. Additionally, to efficiently transfer the knowledge from pretrained full attention models, we utilize continual pretraining (CPT, \citealt{sun2020ernie, mendieta2023towards, yildiz2024investigating, guo2025efficient}) to convert the standard full attention architecture to the proposed \method{}. We perform extensive experiments on twenty-three benchmark datasets covering different context lengths against several competing baselines, and the results demonstrate the effectiveness of \method{}. The contributions of this paper are summarized as follows:
\vspace{-0.5em}
\begin{itemize}[leftmargin=*]
\setlength\itemsep{0em}
\item[\ding{182}] \textit{\textbf{New Perspective}}: We present a dynamic and fine-grained perspective of hybrid transformer models for effective allocation of computation between full attention and its efficient variants.
\item[\ding{183}] \textit{\textbf{Novel Methodology}}: We propose Switch Attention (\method{}), consisting of both the full attention branch and the SWA branch, to adaptively allocate computation for each token's representation at each transformer layer. Additionally, we design an adaptive regularization objective and optimize the model from standard transformers through continual pretraining.
\item[\ding{184}] \textit{\textbf{Empirical Validation}}: We conduct extensive experiments on twenty-three benchmark datasets covering commonsense reasoning, in-context retrieval, and long-context understanding. Further experiments reveal the dynamics of the routing, enabling fine-grained allocation of computation.
\end{itemize}
\section{Related Works}
\begin{figure*}[!t]
    \centering
    \includegraphics[width=\linewidth]{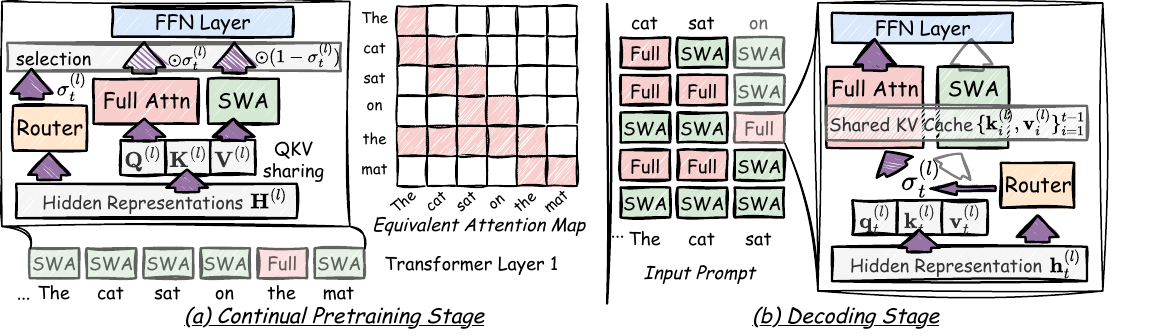}
    \vspace{-4mm}
    \caption{During the continual pretraining stage (a), we compute both the full attention branch and the SWA branch separately. Subsequently, the router is applied to select the attention output from different branches for each token's hidden representation. The selection result is then processed by a feed-forward network (FFN Layer). During the decoding stage (b), the router is used to decide the branch for attention computation. The full attention branch and the SWA branch share a unified KV cache for better efficiency.}
    \vspace{-2mm}
    \label{fig:framework}
\end{figure*}

\subsection{Efficient Attention Mechanism}
The quadratic computational complexity has become a major bottleneck of efficient transformer-based LLMs as input sequences grow longer. This leads to a plethora of efforts in designing efficient attention mechanisms \cite{tay2022efficient, sun2025efficient}. One line of research attempts to reduce the computation of the standard full attention itself by sharing \cite{shazeer2019fast, ainslie2023gqa} and compressing \cite{ge2023model, liu2024deepseek, liu2024minicache, liu2024cachegen} the key and value vectors, or optimizing hardware compatibility \cite{dao2022flashattention, chang2025mosa, sanovar2025leanattention}. Another line of work designs mechanisms to sparsify the standard full attention through heuristic \cite{zaheer2020big, xiao2023efficient} or data-dependent patterns \cite{xu2025xattention, nawrot2025sparse, lai2025flexprefill, yuan2025native}. 

\camrev{Several related methods allocate attention or KV-cache computation at different granularities. DuoAttention \cite{xiao2025duoattention} assigns retrieval or streaming behavior to attention heads, while MoA \cite{fu2024mixture} searches heterogeneous attention spans across heads and layers. FastGen \cite{ge2023model} profiles attention heads and applies head-specific KV-cache compression policies. NSA \cite{yuan2025native} combines compressed global tokens, selected token blocks, and a local window. CCA-Attention \cite{chen2025core} constructs core tokens while preserving local context. MInference \cite{jiang2024minference} selects input-dependent sparse attention patterns. In contrast, \method{} retains exact FullAttn and SWA branches and makes a content-dependent choice for every token at every transformer layer. Recent studies \cite{wu2025retrieval,lin2026retrieval} also suggests that the behavior of attention heads may be dynamic.} 

To circumvent the quadratic complexity, efforts are made in employing sub-quadratic attention \cite{yang2024gated, han2024demystify, dao2024transformers, peng2025rwkv, guo2025log, team2025kimi}. However, their efficiency comes at the expense of degraded long-context performance, as the state matrices limit the memory \cite{guo2025log, nazari2026key, sun2026state}, and in practice, they are often used as an alternative of sliding window attention.

\subsection{Hybrid Transformers}
Hybrid transformer models have been widely adopted to enhance the efficiency of LLMs \cite{Jiang2023Mistral7, team2024gemma, yang2025qwen3}. Many hybrid transformers consist of both standard full attention and its efficient variants, such as sliding window \cite{child2019generating}, sparse \cite{zaheer2020big}, or linear attention \cite{yang2024gated, team2025kimi}. A common practice among existing designs is to interleave full and efficient attention layers (\emph{e.g.}, Figure \ref{fig:motivation}(a)) according to a static ratio empirically defined before the training process \cite{wang2025systematic, openai2025gptoss, xiao2026mimo}. Given the heuristic nature of these methods, efforts are made in automated search for optimal hybrid configurations under certain data distributions \cite{Gu2025JetNemotronEL, acun2025composer, xia2026distill}. Nevertheless, most existing approaches adhere to a static hybrid scheme (using a fixed hybrid pattern for all tokens), either heuristically defined \cite{kitaev2020reformer, team2025kimi, yang2025qwen3} or searched with data \cite{Gu2025JetNemotronEL, liu2025differentiable}, while recent works suggest distinct difficulty in generating tokens \cite{leviathan2023fast, cai2024medusa, gloeckle2024better, liu2024deepseek, fu2025r2r}.

\section{Methodology}
\subsection{Preliminaries and Framework Overview}
\noindent\textbf{Preliminaries and Notations.} We consider the standard language modeling task. During the pretraining stage, the input sentence is first tokenized and then transformed to the hidden representations $\bm H^{(1)}=[\bm h_1^{(1)},\bm h_2^{(1)},...,\bm h_T^{(1)}]\in \mathbb R^{d\times T}$, where $T$ is the sequence length and $d$ is the hidden dimension size. At the $l$-th transformer layer, the input hidden representations $\bm H^{(l)}$ is mapped into queries, keys and values through a linear mapping $\phi$, \emph{i.e.}, $[\bm Q^{(l)},\bm K^{(l)}, \bm V^{(l)}]=\phi(\bm H^{(l)})$. Attention (standard full attention or a more efficient one) is computed using $\bm Q^{(l)},\bm K^{(l)}, \bm V^{(l)}$ to obtain the attended output $\bm O^{(l)}$. Finally, a feed-forward network is applied to obtain the input hidden representations (\emph{i.e.}, $\bm H^{(l+1)}$) of the next transformer layer. The final hidden representation $\bm H^{(L+1)}$ (where $L$ is the number of transformer layers) is used to predict the next token. During the decoding stage, the model generates one token each time with the current query $\bm q_t^{(l)}$, key $\bm k_t^{(l)}$ and value $\bm v_t^{(l)}$, and uses KV cache $\{\bm k_i^{(l)}, \bm v_i^{(l)}\}_{i=1}^{t-1}$ to store previously computed key and value vectors. \footnote{For simplicity, we omit some details here and in the following, such as layer normalization, residual connection, position embeddings, attention heads, \emph{etc.}}

\smallskip
\noindent\textbf{Framework Overview.} The proposed \method{} is a dynamic and fine-grained hybrid of standard full attention and SWA, as illustrated in Figure \ref{fig:framework}. At each transformer layer during the continual pretraining stage, both full attention and SWA are computed using the shared queries, keys and values. A router is applied to the input hidden representations to decide which branch to use for each token. The attended outputs from the two branches are then gathered and processed by a feed-forward layer. An adaptive regularization objective is applied to the output of each router to encourage the model towards the SWA branch. During the decoding stage, the router is first applied to the hidden representation to decide the branch for computing, and the model only computes the selected branch using a shared KV cache. 

\subsection{Dual-branch Routing}
The proposed \method{} consists of two branches: the full attention branch and the SWA branch. Standard full attention excels in capturing long-range dependencies, while SWA efficiently computes attended output using local contexts. Not all tokens require global information at all transformer layers: there is a natural sparsity in both layer \cite{yang2025qwen3, team2025kimi, gao2026hysparse} and token \cite{leviathan2023fast, cai2024medusa, fu2024mixture, fu2025r2r} dimensions. Therefore, we use a router function to decide the specific branch to use.

Concretely, given the input hidden representation $\bm H^{(l)}\in \mathbb R^{d\times T}$ at the $l$-th layer, we first compute the soft gate using the router function $\mathcal R$ as:
\begin{equation}
\label{eq:softgate}
\tilde\sigma^{(l)} = \operatorname{sigmoid}\left(\mathcal R(\bm H^{(l)})\right) \in [0, 1]^T,
\end{equation}
where $\operatorname{sigmoid}(\cdot)$ is the sigmoid function. The (hard) gate indicating the path of computation for each token at the $l$-th layer, $\sigma^{(l)}\in\{0,1\}^T$, is then obtained by applying a threshold $\tau=0.5$, \emph{i.e.},
\begin{equation}
\label{eq:hardgate}
\sigma^{(l)} = \mathds{1}\left(\tilde\sigma^{(l)} > \tau \right) + \tilde\sigma^{(l)} - \operatorname{sg}\left(\tilde\sigma^{(l)}\right),
\end{equation}
where $\mathds{1}(\cdot)$ is an element-wise indicator function and $\operatorname{sg}(\cdot)$ is the stop gradient function. Note that we adopt straight-through estimation (STE, \citealt{fedus2022switch, liu2022nonuniform, huh2023straightening}) to enable differentiation of the router's parameters during training. The gate is used to control whether a corresponding token should be computed with full (when the value is $1$) or sliding window attention (when the value is $0$). Subsequently, we compute both the full attention branch and the SWA branch using a shared set of queries, keys and values, \emph{i.e.},
\begin{align}
\label{eq:qkv}
&[\bm Q^{(l)},\bm K^{(l)}, \bm V^{(l)}]=\phi(\bm H^{(l)}),\\
\label{eq:attn-full}
&\bm O_{\text{FULL}}^{(l)} = \mathcal A_{\text{FULL}}^{(l)}(\bm Q^{(l)},\bm K^{(l)}, \bm V^{(l)}),\\
\label{eq:attn-swa}
&\bm O_{\text{SWA}}^{(l)} = \mathcal A_{\text{SWA}}^{(l)}(\bm Q^{(l)},\bm K^{(l)}, \bm V^{(l)}),
\end{align}
where $\mathcal A_{\text{FULL}}^{(l)}$ and $\mathcal A_{\text{SWA}}^{(l)}$ denote full attention and SWA, respectively. The gate is then used to select the output of attention $\bm O^{(l)}$, \emph{i.e.},
\begin{equation}
\label{eq:attnout}
\bm O^{(l)} = \sigma^{(l)} \odot \bm O_{\text{FULL}}^{(l)} + (1 - \sigma^{(l)}) \odot \bm O_{\text{SWA}}^{(l)},
\end{equation}
where $\odot$ denotes element-wise product (in the sequence length dimension). Finally, the feed-forward network (FFN) is applied to obtain the hidden representation at the $(l+1)$-th layer. 

During inference, the attention output is computed similarly. At the prefill stage, the attention output can be computed via Eq.\ref{eq:softgate}-\ref{eq:attnout}, and especially, the computation of the two branches could be further merged with a hardware-efficient implementation. At the decode stage, we compute the (hard) gate for the $t$-th token at the $l$-th layer as a scaler $\sigma^{(l)}_t \in \{0, 1\}$, and then compute the full attention or SWA accordingly. Notably, in the decoding stage, both branches share a single set of KV-cache to increase inference efficiency. We summarize the inference procedure of \method{} in Appendix \ref{app:inference}.

\subsection{Adaptive Regularization}
In the dual-branch routing framework, the full attention branch provides richer information than the SWA branch. Therefore, to encourage the model towards the SWA branch for efficiency, we also design an adaptive regularization objective that is optimized during continual pretraining. Conceivably, when the ground-truth token exhibits lower log-likelihood under the predicted distribution, the model may struggle to predict this token, and therefore, we expect higher ratios for the more powerful full attention branch. Additionally, when the two branches disagree, we also tend toward the more reliable full attention branch. These require the regularization objective to be more adaptive.

Specifically, given the soft gate of the $t$-th token at the $l$-th transformer layer $\tilde\sigma_t^{(l)}$, the regularization objective is designed as follows:
\begin{equation}
\label{eq:ada-reg}
\mathcal L_t^{(l)} = \gamma_t^{(l)} \operatorname{sp}\left(\mathcal R(\bm h_t^{(l)})\right),
\end{equation}
where $\gamma_t^{(l)}$ is the adaptive weight of regularization discussed in the following, and $\operatorname{sp}(x)=\ln(1+e^x)$ is the softplus function, a smooth version of the ReLU function \cite{nair2010rectified}, providing a small penalty to values less than $0$ and increasing penalty to values greater than $0$. The adaptive weight $\gamma_t^{(l)}$ of the $t$-th token at the $l$-th layer is determined by several factors: (1) the base weight $\gamma_{\text{base}}$, (2) the negative log-likelihood (NLL) of the ground-truth token under the currently predicted distribution of language modeling, \emph{i.e.}, $\operatorname{NLL}(t)$, and (3) the disagreement of the two branches as measured by the mean squared error (MSE), \emph{i.e.}, $\norm{\bm o_{\text{FULL}}^{(t,l)}-\bm o_{\text{SWA}}^{(t,l)}}^2_2$, where $\bm o_{\text{FULL}}^{(t,l)}$ and $\bm o_{\text{SWA}}^{(t,l)}$ denotes the output of the full attention branch and the SWA branch of the $t$-th token at the $l$-th layer. Considering these factors, the adaptive weight $\gamma_t^{(l)}$ can be written as follows:
\begin{equation}
\label{eq:ada-weight}
\gamma_t^{(l)} =  \frac{\gamma_{\text{base}}}{\varepsilon + \operatorname{NLL}(t) + \alpha \cdot \norm{\bm o_{\text{FULL}}^{(t,l)}-\bm o_{\text{SWA}}^{(t,l)}}^2_2},
\end{equation}
where $\varepsilon$ and $\alpha$ are hyper-parameters. 

\camrev{The NLL and branch MSE are used only to construct the adaptive regularization weight during teacher-forced CPT. At inference, the routing decision depends only on $\mathcal R(\bm H)$, and the model computes only the selected attention branch.}

\smallskip
\noindent\textbf{Remark.} Eq.\ref{eq:ada-weight} adaptively \camrev{reweights} the regularization objective of the router. The hyperparameter $\varepsilon$ avoids division by zero and provides an upper bound of the weight, \emph{i.e.}, $\gamma_t^{(l)}\le \gamma_{\text{base}}\cdot \varepsilon^{-1}$. The NLL and MSE terms reduce the weight of regularization when the token prediction loss increases (high NLL) and/or the two branches diverge (high MSE). Under such conditions, the reduced regularization weights encourage the router to use fewer SWA branches and more full attention branches, since the SWA branch may be less reliable. \camrev{Although the full-attention ratio emerges from training, it can be steered toward a target compute budget by adjusting $\gamma_{\text{base}}$ according to the observed routing ratio: increasing $\gamma_{\text{base}}$ encourages more SWA decisions, whereas decreasing it relaxes the efficiency-oriented penalty. The experiments in this work use a fixed $\gamma_{\text{base}}$. Further motivation for this design is provided in Appendix \ref{app:adareg}.}

\begin{algorithm}[tb]
    \caption{The continual pretraining of \method{}}
    \label{alg:main}
    \textbf{Requires}: The pretrained standard full attention transformer model $\mathcal M^{pre}$, hyperparameter $\varepsilon$ and $\alpha$, number of continual pretraining iterations $N$.\\
    \textbf{Ensures}: The continual pretrained \method{} $\mathcal M$.
    
    \begin{algorithmic}[1] 
    \STATE Initialize the weights of $\mathcal A_\text{FULL}^{(l)}$, $\mathcal A_\text{SWA}^{(l)}$ in $\mathcal M$ using the weights $\tilde{\mathcal A}_\text{FULL}^{(l)}$ from $\mathcal M^{pre}$;
    \FOR{Iterations $i=1,2,...,N$}
    \STATE Tokenize the training batch of the $i$-th iteration and transform them into initial hidden representations $\bm H^{(1)}$;
    \FOR{Layers $l=1,2,...,L$}
    \STATE Compute the soft gate $\tilde\sigma^{(l)}$ using Eq.\ref{eq:softgate};
    \STATE Obtain the hard gate $\sigma^{(l)}$ using threshold $\tau$ according to Eq.\ref{eq:hardgate};
    \STATE Compute the routed output $\bm O^{(l)}$ using Eq.\ref{eq:qkv}-\ref{eq:attnout} and apply the FFN layer;
    \STATE Compute the router logits $\mathcal R(\bm h_t^{(l)})$ and MSE $||\bm o_{\text{FULL}}^{(t,l)}-\bm o_{\text{SWA}}^{(t,l)}||^2_2$ for each token;
    \ENDFOR
    \STATE Compute the NLL for each token as $\operatorname{NLL}(t)$;
    \vspace{-4mm}
    \STATE Compute the adaptive weight $\gamma_t^{(l)}$ using Eq.\ref{eq:ada-weight} to obtain adaptive regularization $\mathcal L_t^{(l)}$;
    \STATE Compute the language modeling loss $\mathcal L_\text{LM}$ and the final loss $\mathcal L$ using Eq.\ref{eq:final-loss};
    \ENDFOR
    \STATE Update the parameters of $\mathcal M$ using $\nabla \mathcal L$;
    \end{algorithmic}
\end{algorithm}

\subsection{Continual Pretraining}
\label{sec:cpt}
Different from many hybrid transformers that require training from scratch, the proposed \method{} adopts a continual pretraining approach, utilizing the learned knowledge from an existing off-the-shelf full attention transformer.

In particular, given a standard transformer and its full attention module at the $l$-th layer $\tilde {\mathcal A}_{\text{FULL}}^{(l)}$, we initialize parameters of both the full attention branch ${\mathcal A}_{\text{FULL}}^{(l)}$ and the SWA branch ${\mathcal A}_{\text{SWA}}^{(l)}$ of \method{} using the parameters of $\tilde {\mathcal A}_{\text{FULL}}^{(l)}$. Other parameters, like the FFN layer, are also inherited from the standard pretrained transformer. During the continual pretraining stage, we train the entire hybrid transformer using both the language modeling objective $\mathcal L_\text{LM}$ and the aforementioned adaptive regularization objective, \emph{i.e.},
\begin{equation}
\label{eq:final-loss}
\mathcal L = \mathcal L_\text{LM} + \frac{1}{LT}\sum_{l=1}^L\sum_{t=1}^T \mathcal L_t^{(l)}.
\end{equation}
By continual pretraining from an existing standard transformer, we save computation resources and stabilize the optimization process. The continual pretraining process of the proposed \method{} is presented in Algorithm \ref{alg:main}.


\subsection{Hardware Compatibility}
\begin{camrevblock}
During CPT, \method{} computes both full attention and SWA. The SWA
branch can be implemented efficiently using FlashAttention \cite{dao2022flashattention}. The router maps each hidden representation to a scalar and can be fused
into the QKV projection as one additional output, making its overhead negligible relative to attention.

During inference, efficient attention does not automatically translate into efficient hardware execution. For example, in Q-outer sparse attention under GQA, \citet{lai2026minimax} derive its BF16 arithmetic intensity. Using $G$ for the GQA group size and $S$ for the number of KV tokens processed by a query, its arithmetic intensity can be written as
\begin{equation}
\label{eq:ai-qouter}
\mathrm{AI}_{\mathrm{Q\text{-}outer}}
=
\frac{GS}{G+S}
\approx G,
\qquad S\gg G.
\end{equation}
Thus, the arithmetic intensity is largely determined by the GQA group size, and many Q-outer sparse-attention kernels require a sufficiently large $G$ for efficient execution \cite{yuan2025native,liu2025deepseek,zeng2026glm}.

\method{} instead groups routed query positions into branch-specific, position-sorted index lists and gathers them in tiles of $B_q$ positions for each query head. For the $G$ query heads sharing one KV head, the kernel processes each query head in a separate thread block, so both computation and the issued KV loads scale with $G$ under the same IO accounting. For a query tile traversing $S$ KV tokens, its dominant BF16 arithmetic intensity, ignoring lower-order traffic, is
\begin{equation}
\label{eq:ai-swiattn}
\begin{aligned}
\mathrm{AI}_{\method}
&=
\frac{4GB_qSd_h}
     {4GB_qd_h+4GSd_h}
=
\frac{B_qS}
     {B_q+S} \\
&\approx B_q,
\qquad S\gg B_q.
\end{aligned}
\end{equation}
Thus, KV reuse is determined by the kernel-controlled query-tile size $B_q$, rather than the GQA group size $G$. \method{} consequently avoids the large-group requirement common to many sparse-attention methods \cite{yuan2025native,liu2025deepseek,zeng2026glm}. This makes \method{} an efficient and hardware-friendly attention mechanism under smaller group sizes. An exemplar kernel implementation is provided in Appendix~\ref{app:kernel}.

During decoding, \method{} remains compatible with existing
optimization infrastructure, including PagedAttention \cite{kwon2023efficient} and Split-KV strategies such as FlashInfer \cite{ye2025flashinfer}.
\end{camrevblock}

\begin{table*}[!t]
\centering
\caption{Performance on common sense reasoning datasets. Best results are in \textbf{bold} and second-best \underline{underline}.}
\label{tab:csresults}
\vspace{-2mm}
\resizebox{\textwidth}{!}{%
\def\arraystretch{1.2}
\renewcommand\tabcolsep{5pt}
\begin{tabular}{
  >{\raggedright\arraybackslash}p{2cm}
  !{\vrule width 1.0pt}
  *{2}{>{\centering\arraybackslash}p{1.1cm}}
  >{\centering\arraybackslash}p{1.1cm}
  !{\vrule width 1.0pt}
  *{8}{>{\centering\arraybackslash}p{1.1cm}}
  >{\centering\arraybackslash}p{1.1cm}
}
\Xhline{1.3pt}
\rowcolor{tableheadcolor!50} Datasets & LMB. & Wiki. & AVG & BoolQ & PIQA & SIQA & Hella. & Wino. & ARC-e & ARC-c & OBQA & AVG \\
\Xhline{1.0pt}
\rowcolor{tableheadcolor!50} Metrics
& ppl$\downarrow$ & ppl$\downarrow$ & ppl$\downarrow$ & acc$\uparrow$ & acc$\uparrow$ & acc$\uparrow$ & acc\_n$\uparrow$ & acc$\uparrow$ & acc$\uparrow$ & acc\_n$\uparrow$ & acc\_n$\uparrow$ & acc$\uparrow$ \\
\Xhline{1.3pt}
SWA-ZS & \underline{7.16} & 16.6 & 11.9 & \underline{62.2} & \underline{74.0} & 41.4 & \underline{60.1} & 57.1 & \underline{62.9} & \underline{34.3} & 35.4 & 50.5 \\
\rowcolor{tablerowcolor}SWA-CPT & \textbf{7.02} & {15.6} & \underline{11.3} & {62.1} & \textbf{74.5} & \underline{42.0} & \underline{60.1} & \textbf{57.9} & 62.7 & \textbf{35.5} & 35.2 & 52.6 \\
\camrev{SH (3:1)} & 7.32 & \underline{15.2} & \underline{11.3} & \textbf{62.3} & {73.7} & {\underline{42.0}} & \textbf{60.2} & \textbf{57.9} & 62.3 & 33.6 & \underline{36.7} & \underline{53.6} \\
\rowcolor{tablerowcolor}FullAttn & \underline{7.16} & \textbf{15.0} & \textbf{11.1} & \underline{62.2} & \underline{74.0} & 41.4 & \underline{60.1} & 57.1 & \underline{62.9} & \underline{34.3} & 35.4 & 53.4 \\
\Xhline{1.0pt}
\rowcolor{ourrowname!50}\method{} & 7.28 & \textbf{15.0} & \textbf{11.1} & 62.0 & \underline{74.0} & \textbf{42.3} & \textbf{60.2} & \underline{57.8} & \textbf{63.0} & 34.0 & \textbf{36.8} & \textbf{53.7} \\
\Xhline{1.3pt}
\end{tabular}}
\vspace{-1mm}
\vspace{-3mm}
\end{table*}

\begin{table}[!t]
\centering
\caption{Performance on in-context retrieval datasets. Best results are in \textbf{bold} and second-best \underline{underline}.}
\label{tab:riresults}
\vspace{-2mm}
\resizebox{\linewidth}{!}{%
\def\arraystretch{1.2}
\renewcommand\tabcolsep{3pt}
\begin{tabular}{
  >{\raggedright\arraybackslash}p{1.9cm}
  *{1}{>{\centering\arraybackslash}p{1.1cm}}
  *{6}{>{\centering\arraybackslash}p{0.9cm}}
}
\Xhline{1.3pt}
\rowcolor{tableheadcolor!50} Datasets & SWDE & FDA & TQA & NQ & {\small DROP} & SQD & AVG \\
\Xhline{1.3pt}
SWA-ZS       & 52.0 & 22.5 & 64.6 & 30.9 & 31.1 & 49.2 & 41.7 \\
\rowcolor{tablerowcolor}SWA-CPT   & 49.1 & 24.0 & \textbf{65.7} & 28.4 & \underline{32.2} & \underline{49.7} & 41.5 \\
\camrev{SH (3:1)} & 56.5 & 67.9 & 65.1 & 32.7 & 31.3 & \textbf{50.1} & \underline{50.6} \\
\rowcolor{tablerowcolor}FullAttn  & \textbf{60.8} & \textbf{76.4} & 64.6 & \underline{35.1} & 31.1 & 49.2 & \textbf{52.9} \\
\Xhline{1.0pt}
\rowcolor{ourrowname!50}\method{} & \underline{60.2} & \underline{75.1} & \underline{65.2} & \textbf{35.2} & \textbf{32.4} & 49.4 & \textbf{52.9} \\
\Xhline{1.3pt}
\end{tabular}}
\vspace{-3mm}
\end{table}

\section{Experiments}
\subsection{Experimental Setups}
\noindent\textbf{Datasets.} Three types of datasets are used. \emph{$\blacktriangleright$ Commonsense Reasoning}, including Lambada (LMB., \citealt{paperno2016lambada}), Wikitext (Wiki., \citealt{merity2016pointer}), BoolQ \cite{clark2019boolq}, PIQA \cite{bisk2020piqa}, SIQA \cite{sap2019social}, HellaSwag (Hella., \citealt{zellers2019hellaswag}), WinoGrande (Wino., \citealt{sakaguchi2021winogrande}), ARC-easy and ARC-challenge (ARC-e and ARC-c, \citealt{clark2018think}), and OpenBookQA (OBQA, \citealt{mihaylov2018can}); \emph{$\blacktriangleright$ In-Context Retrieval} from \citealt{arora2024just}, including SWDE \cite{lockard2019openceres}, FDA \cite{arora2023language}, TriviaQA (TQA, \citealt{joshi2017triviaqa}), NQ \cite{kwiatkowski2019natural}, DROP \cite{dua2019drop}, and SQuAD (SQD, \citealt{rajpurkar2018know}); \emph{$\blacktriangleright$ Long-Context Understanding} from LongBench-E \cite{bai2024longbench}, including MultiFieldQA (MFQA, \citealt{bai2024longbench}), TRec \cite{li2002learning}, MultiNews (MNs, \citealt{fabbri2019multi}), GovReport (GvR, \citealt{huang2021efficient}), Qasper (Qas, \citealt{dasigi2021dataset}), TriviaQA (TQA, \citealt{joshi2017triviaqa}), and SAMSum (SSM, \citealt{gliwa2019samsum}). More details about the datasets are provided in Appendix \ref{app:dataset}.

\smallskip
\noindent\textbf{Baselines.} \camrev{We compare the proposed \method{} against several baselines: sliding window attention in zero-shot and continual-pretraining forms, denoted SWA-ZS and SWA-CPT; static hybrids of full attention and SWA commonly used in larger industrial LLMs \cite{brown2020language, openai2025gptoss, xiao2026mimo}, denoted SH (3:1) and SH (6:1) according to their SWA-to-FullAttn layer ratios; DuoAttention \cite{xiao2025duoattention}; and full attention, denoted FullAttn. More details about the baseline methods are provided in Appendix \ref{app:baseline}.}

\smallskip
\noindent\textbf{Implementation Details.} For the transformer architecture, we use a $1.5$B language model with $22$ transformer layers (\emph{i.e.}, $L=22$). Each layer contains a dual-branch routing module and an FFN layer. The router function $\mathcal R(\cdot)$ is implemented with a linear mapping, and the threshold $\tau$ is set to $0.5$. For the SWA branch, the window size is set to $1024$. In adaptive regularization, we set the base weight $\gamma_\text{base}$ to $1\times 10^{-3}$, the $\varepsilon$ to $0.1$ and the weight $\alpha$ to $100$. For model optimization, we first pretrain the $1.5$B full attention model on $4$K-length sequences using $240$B tokens and then continually pretrain the \method{} model for $20$B tokens. To evaluate tasks of longer sequences, we optimize the $32$K full attention model by pretraining on $4$K context lengths for $192$B tokens and then extending the context length using YaRN \cite{peng2023yarn} to $32$K with $24$B tokens. The corresponding \method{} is obtained by continual pretraining on the full attention model for another $8$B tokens. More details can be found in Appendix \ref{app:implementation}.

\subsection{Main Results}
We compare the performance of \method{} against baselines on $\camrev{23}$ benchmark datasets, and the results are shown in Table \ref{tab:csresults}, Table \ref{tab:riresults} and Table \ref{tab:lbresults}. 

\smallskip
\noindent\textbf{Commonsense Reasoning (Table \ref{tab:csresults}).} As the contexts of most instances are within the sliding window size (\emph{i.e.}, $1024$), the performance of \method{} is similar to the baselines, including both full attention and SWA models, \rev{demonstrating that \textit{\textbf{Obs.\ding{182} \method{} maintains the commonsense knowledge of full attention after CPT.}}}

\smallskip
\noindent\textbf{In-Context Retrieval (Table \ref{tab:riresults}).} \camrev{For these datasets, the contexts are within the $4096$ token limit of $4$K-length models, and some instances require attention spans beyond the sliding window. \textit{\textbf{Obs.\ding{183} \method{} matches FullAttn while outperforming SH (3:1) and the SWA baselines.}} This corresponds to a relative improvement of $4.5\%$ over SH (3:1).}

\smallskip
\noindent\textbf{Long-Context Understanding (Table \ref{tab:lbresults}).} \camrev{For these datasets, the contexts of most instances grow beyond the $1024$ window size and can be as long as $32$K, which requires strong long-context capability. \textit{\textbf{Obs.\ding{184} \method{} matches FullAttn on average while outperforming DuoAttention, both StaticHybrid configurations, and the SWA baselines.}} Compared with SH (3:1), \method{} achieves a relative improvement of $6.3\%$, despite using full attention less frequently (as shown in Section \ref{sec:further-analysis}).}

\camrev{Overall, \method{} achieves performance parity with FullAttn while improving over the other baselines on tasks that require longer contexts. Results on 3B models are in Appendix \ref{app:3b}.}

\begin{table}[t]
\centering
\caption{Performance on long-context understanding. Best results are in \textbf{bold} and second-best \underline{underline}.}
\label{tab:lbresults}
\vspace{-2mm}
\resizebox{\columnwidth}{!}{%
\def\arraystretch{1.2}
\renewcommand\tabcolsep{1.5pt}
\begin{tabular}{
  >{\raggedright\arraybackslash}p{2.2cm}
  *{8}{>{\centering\arraybackslash}p{1.1cm}}
}
\Xhline{1.3pt}
\rowcolor{tableheadcolor!50} Datasets & GvR & MNs & MFQA & Qas & TQA & SSM & TRec & AVG \\
\Xhline{1.3pt}
SWA-ZS & 19.2 & 18.2 & 18.5 & 10.5 & 51.1 & 27.4 & 18.7 & 23.4  \\
\rowcolor{tablerowcolor}SWA-CPT & 19.2 & 18.7 & 24.2 & 13.1 & 57.8 & 30.7 & 53.0 & 31.0 \\
\camrev{DuoAttention} & \camrev{21.6} & \camrev{19.9} & \camrev{30.9} & \camrev{14.5} & \camrev{62.4} & \camrev{29.4} & \camrev{58.7} & \camrev{33.9} \\
\rowcolor{tablerowcolor}\camrev{SH (6:1)} & \camrev{21.0} & \camrev{19.0} & \camrev{26.2} & \camrev{\textbf{20.1}} & \camrev{60.0} & \camrev{31.2} & \camrev{55.7} & \camrev{33.3} \\
\camrev{SH (3:1)} & 22.6 & 20.0 & 29.6 & 14.5 & 68.1 & \underline{32.3} & 59.3 & \underline{35.2} \\
\rowcolor{tablerowcolor}FullAttn & \underline{24.9} & \underline{20.9} & \underline{33.9} & \camrev{\underline{18.6}} & \underline{69.3} & 30.2 & \textbf{63.7} & \textbf{37.4} \\
\Xhline{1.0pt}
\rowcolor{ourrowname!50}\method{} & \textbf{25.0} & \textbf{21.3} & \textbf{35.3} & {18.0} & \textbf{69.9} & \textbf{32.4} & \underline{60.0} & \textbf{37.4} \\
\Xhline{1.3pt}
\end{tabular}}
\vspace{-1mm}
\vspace{-3mm}
\end{table}

\subsection{Ablation Studies}
To investigate the effect of different mechanisms in the proposed method, we design several variants of the original \method{}. Specifically, V1 disables adaptive regularization and uses $\gamma_\text{base}$ instead. V2 removes the NLL term in Eq.\ref{eq:ada-weight}. V3, V4, and V5 changes $\alpha$ from $100$ to $10$, $1$, and $0$, respectively. V6 and V7 perturb $\varepsilon$ from $0.1$ to $0.2$ and $0.01$. We present results on three summarization datasets from LongBench-E in Table \ref{tab:ablation}.

\smallskip
\noindent\textbf{Effect of Adaptive Regularization.} From V1, V2, and V5, we can see that \textit{\textbf{Obs.\ding{185} adaptive regularization, along with several factors in adaptive weights, is {effective}}}. The removal of adaptive regularization reduces average performance by $1.3$. Besides, when the NLL (V2) or MSE (V5) terms are removed, the performance also declines.

\smallskip
\noindent\textbf{Effects of Hyperparameters.} V3 and V4 show the effect of the hyperparameter of $\alpha$. When $\alpha$ is changed, the accuracy drops, as $\alpha$ balances the relative importance of the MSE term in adaptive regularization. Variants V6 and V7 show the effect of $\varepsilon$. The model is less sensitive to this hyperparameter, and when $\varepsilon$ is perturbed, the accuracy decreases slightly. $\varepsilon$ controls the upper bound of the adaptive weights, which affects the optimization process.

\begin{table}[t]
\centering
\caption{Ablation studies of \method{}.}
\label{tab:ablation}
\vspace{-2mm}
\resizebox{\linewidth}{!}{%
\def\arraystretch{1.2}
\renewcommand\tabcolsep{5pt}
\begin{tabular}{
  >{\raggedright\arraybackslash}p{3.5cm}
  *{3}{>{\centering\arraybackslash}p{1.1cm}}
  *{1}{>{\centering\arraybackslash}p{1.1cm}}
}
\Xhline{1.3pt}
\rowcolor{tableheadcolor!50} Variants & GvR & MNs & SSM & AVG  \\
\Xhline{1.3pt}
V1: \emph{w/o} Adaptive Reg.       & 24.1  & 20.7 & 30.0 & 24.9  \\
\rowcolor{tablerowcolor}V2: \emph{w/o} the NLL term   & 24.3 & 20.9 & 30.7 & 25.3  \\
\Xhline{1.0pt}
V3: $\alpha=10$ & 24.1 & 20.9 & 30.5 & 25.2  \\
\rowcolor{tablerowcolor}V4: $\alpha=1$  & 23.8  & 20.9 & 29.9 & 24.9  \\
V5: $\alpha=0$       & 23.9  & 20.6 & 29.5 & 24.7  \\
\Xhline{1.0pt}
\rowcolor{tablerowcolor}V6: $\varepsilon=0.2$   & 24.7 & 20.9 & 31.7  & 25.8  \\
V7: $\varepsilon=0.01$ & 24.4 & 20.8 & 30.1  &  25.1 \\
\Xhline{1.0pt}
\rowcolor{ourrowname!50}\method{} & 25.0 & 21.3 & 32.4 & 26.2  \\
\Xhline{1.3pt}
\end{tabular}}
\vspace{-3mm}
\end{table}

\subsection{Further Analysis}
\label{sec:further-analysis}
\begin{figure}[t]
    \centering
    \includegraphics[width=\linewidth]{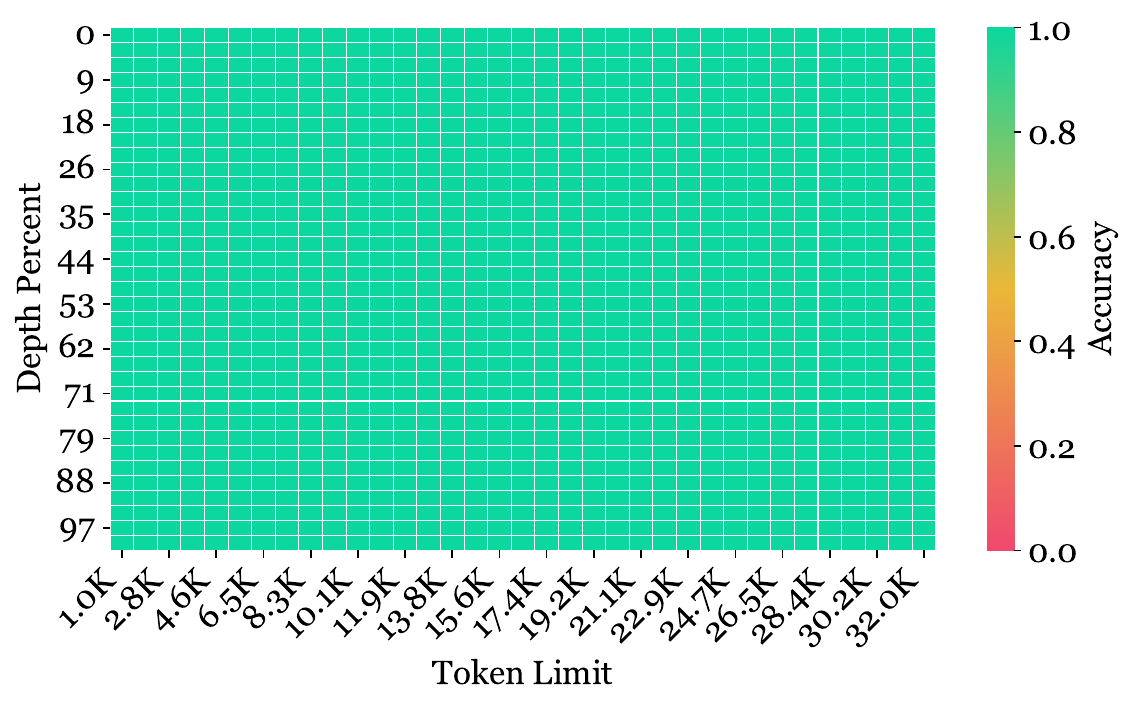}
    \vspace{-7mm}
    \caption{Needle-in-a-haystack retrieval accuracy in different context positions (depth percentage) across $32$K context lengths. The proposed \method{} achieves perfect retrieval accuracy with its dynamic and fine-grained hybrid attention mechanism.}
    \label{fig:niah}
    \vspace{-1mm}
\end{figure}

\noindent\textbf{Needle-in-a-Haystack.} We also provide experimental results on the commonly-used Needle-In-A-Haystack (NIAH) task using RULER \cite{hsieh2024ruler}. Specifically, we test $35$ different context positions (\emph{i.e.}, depth percent) and $35$ context lengths (\emph{i.e.}, token limit). The retrieval tests are repeated three times for each case, and the results are visualized in Figure \ref{fig:niah}. As can be seen from the results, \textit{\textbf{Obs.\ding{186} \method{} achieves {perfect} retrieval accuracy across various scenarios}}, showing that \method{} maintains the strong retrieval capability of full attention with its dynamic routing mechanism models under long context lengths.

\begin{figure}[t]
    \centering
    
    \begin{subfigure}[b]{0.6\linewidth}
        \vspace{0pt}
        \centering
        \includegraphics[width=\linewidth]{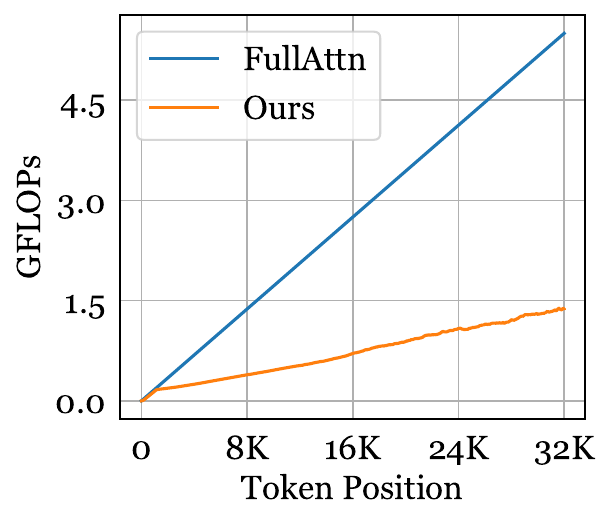}
        \captionsetup{skip=5pt}
        \caption{Prefill}
        \label{fig:eff-prefill}
    \end{subfigure}
    \hfill
    \begin{subtable}[b]{0.36\linewidth}
        \vspace{0pt}
        \centering
        \resizebox{\linewidth}{!}{%
        \begin{tabular}{lcc}
            \Xhline{1.3pt}
            \rowcolor{tableheadcolor!50} & \multicolumn{2}{c}{Mem. Access} \\
            \cline{2-3} \rowcolor{tableheadcolor!50} \multirow{-2}{*}{Pos.} & Ours & Full \\
            \Xhline{1.0pt}
            \rowcolor{tablerowcolor}\multicolumn{3}{l}{\textit{Standard}}\\
            \Xhline{1.0pt}
            8K & 2.3K & 8K \\
            \rowcolor{tablerowcolor}16K & 4.2K & 16K \\
            24K & 6.2K & 24K \\
            \rowcolor{tablerowcolor}32K & 7.3K & 32K \\
            \Xhline{1.0pt}
            \multicolumn{3}{l}{\textit{Simple}}\\
            \Xhline{1.0pt}
            \rowcolor{tablerowcolor}16K & 1.7K & 16K \\
            32K & 2.6K & 32K \\
            \Xhline{1.3pt}
        \end{tabular}}
        \caption{Decode}
        \label{tab:eff-decode}
    \end{subtable}
    \vspace{-1mm}
    \caption{Efficiency comparison under different token positions of \method{} (ours) and FullAttn baseline. We report GFLOPs for the prefill stage and memory access in terms of averaged tokens in the decode stage.}
\end{figure}

\smallskip
\noindent\textbf{Inference Efficiency.} We compare the inference efficiency of \method{} and the FullAttn baseline. For the prefill stage, we plot the amount of computation required (in terms of GFLOPs) at every token position, following \citealt{liu2025deepseek}, and the results are shown in Figure \ref{fig:eff-prefill}. For the decode stage, we measure efficiency with the amount of memory access of the KV cache (in terms of the number of tokens at each layer on average), as it is the primary bottleneck of the decoding speed of attention. As the efficiency of \method{} is related to the data distribution, we provide results on both standard (under the same distribution of training corpus) and simple (under the distribution of needle-in-a-haystack task where the needle is within the sliding window) setups in Figure \ref{tab:eff-decode}. \camrev{At a $32$K context length, \method{} reduces KV-cache access by approximately $4.4\times$ under the standard data distribution and by up to $12\times$ under the simple data distribution.}

\begin{figure}[t]
    \centering
    \includegraphics[width=\linewidth]{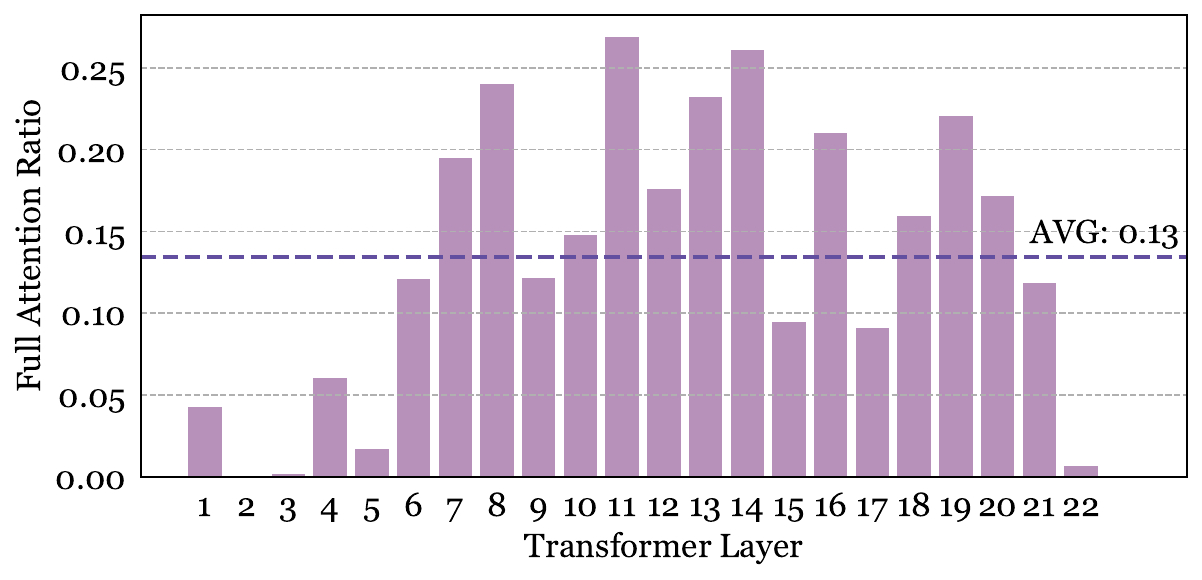}
    \vspace{-7mm}
    \caption{The ratio of using full attention branches across different transformer layers. The averaged full attention ratio is around $0.13$.}
    \vspace{-1mm}
    \label{fig:fr-bar}
\end{figure}

\smallskip
\noindent\textbf{Ratios of Full Attention across Layers.} We then investigate how different transformer layers behave in dual-branch routing. Specifically, we collect the routing decisions of each transformer layer and calculate the ratio of the model choosing full attention under the $32$K continual pretraining data distribution and count the decisions of $400$M tokens. The results are presented in Figure \ref{fig:fr-bar}. As can be seen from the results, \textit{\textbf{Obs.\ding{188} different transformer layers exhibit distinct patterns of routing}}: some layers (\emph{e.g.}, Layer 1, 2, 3, 22) tend to use more SWA branches while others (\emph{e.g.},  Layer 8, 11, 14) use full attention branches more frequently. This suggests different layers specialize in different functionalities. Additionally, we also compute the averaged full attention ratio of all layers, and the result is around $0.13$, which is lower than the $0.25$ ratio of conventional static hybrid models with one full attention layer after three SWA layers.

\begin{figure}[t]
    \centering
    \includegraphics[width=\linewidth]{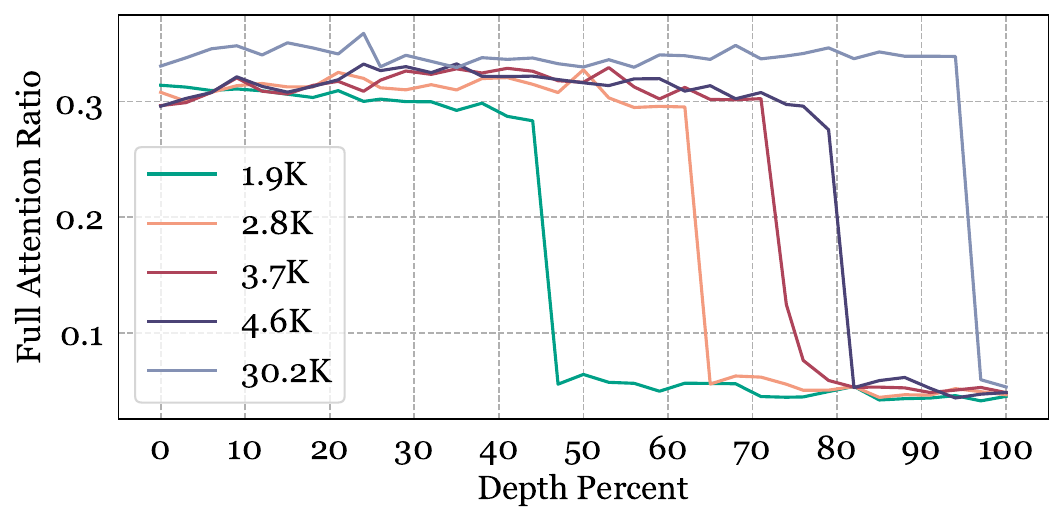}
    \vspace{-5mm}
    \caption{The ratio of full attention branches \textit{when generating the needle} in the needle-in-a-haystack retrieval task. When the needles fall within the sliding window, the full attention ratio is low, while a higher ratio is observed when the needle is outside the sliding window.}
    \vspace{-1mm}
    \label{fig:niah-line}
\end{figure}

\smallskip
\noindent\textbf{Dynamics in Dual-Branch Routing.} We also investigate the dynamics of dual-branch routing. When the next token can be easily predicted based on information within the sliding window, we expect the model to use the SWA branch more frequently, and when the next token is hard to forecast with local information, the full attention branch should be more frequently used as a default choice. To verify whether \method{} behaves in a computation efficient way, we resort to the NIAH retrieval task, where needles to be retrieved are inserted in different locations of the context (\emph{i.e.}, different depth percent, higher percents indicate closer distance to the end of the prompt), and measure the full attention ratio of \method{} when generating the answer (\emph{i.e.}, the inserted needle). The results are presented in Figure \ref{fig:niah-line}, where we illustrate the ratio under different context lengths (\emph{i.e.}, $1.9$K, $2.8$K, $3.7$K, $4.6$K, $30.2$K). The results show that when the needle is inserted in lower depth percentages (outside the sliding window), the model chooses around $30$\% of full attention, while when the model is closer to the end of the prompt (within the sliding window), a lower ratio of full attention around $5$\% is observed. This demonstrates that \textit{\textbf{Obs.\ding{189} \method{} can dynamically and efficiently decide the specific branch to use based on the context}}.

\section{Conclusion}
This paper presents Switch Attention (\method{}), a novel hybrid transformer that can dynamically route the hidden representation to either full attention or sliding window attention at each transformer layer, enabling efficient and fine-grained allocation of computation based on the contexts. \method{} can be optimized via continual pretraining from off-the-shelf standard full attention transformers. An adaptive regularization objective is designed to enhance efficiency while maintaining accuracy. Extensive experiments demonstrate the effectiveness of the proposed \method{}, and further empirical results reveal its dynamics and efficiency.

\section*{Limitations}
\camrev{Our experiments cover models up to $3$B parameters, and the 3B model uses the same hyperparameter configuration as the 1.5B model without a scale-specific search. Effectiveness at larger industrial scales therefore remains to be verified. In addition, the efficiency of \method{} depends on the routing distribution induced by the input. In the worst case, where every token at every layer selects FullAttn, \method{} has the same attention and KV-cache access complexity as FullAttn together with the small router cost, and therefore provides no speedup.}

\section*{Acknowledgements}
This paper is partially supported by the National Key Research and Development Program of China with Grant No. 2023YFC3341203 as well as the National Natural Science Foundation of China with Grant Number 62276002.


\bibliography{7_ref}

\appendix

\clearpage
\section{Inference Procedure of \method{}}
\label{app:inference}
We provide the inference procedure of the proposed \method{} in Algorithm \ref{alg:inference}.
\begin{algorithm}[tb]
    \caption{The inference procedure of \method{}}
    \label{alg:inference}
    \textbf{Requires}: The \method{} model $\mathcal M$, the prompt $P$.\\
    \textbf{Ensures}: The response sequence $R$.
    
    \begin{algorithmic}[1] 
    \STATE Tokenize $P$ and convert it to the initial hidden representations $\bm H^{(1)}$;
    \FOR{Layer $l=1,2,...,L$}
    \STATE Compute $\bm Q^{(l)},\bm K^{(l)}, \bm V^{(l)}$ and initialize the shared KV-cache $\mathcal C^{(l)}=\{\bm k_i^{(l)}, \bm v_i^{(l)}\}_{i=1}^{t-1}$;
    \STATE Compute the gate (Eq.\ref{eq:softgate}-\ref{eq:hardgate}), attention outputs (Eq.\ref{eq:attn-full}-\ref{eq:attnout}), and apply the FFN layer.
    \ENDFOR
    \STATE Generate the first token and append to $R$;
    \WHILE{not reaching EOS token}
    \FOR{Layer $l=1,2,...,L$}
    \STATE Compute $\bm q_t^{(l)}, \bm k_t^{(l)}, \bm v_t^{(l)}$ and update $\mathcal C^{(l)}$;
    \STATE Compute the gate $\sigma^{(l)}_t$ using Eq.\ref{eq:softgate}-\ref{eq:hardgate};
    \IF{$\sigma^{(l)}_t=1$}
    \STATE Compute full attention output $\bm o^{(l,t)}=\bm o^{(l,t)}_\text{FULL}=\mathcal A^{(l)}_\text{FULL}(\bm q_t^{(l)}, \mathcal C^{(l)})$;
    \ELSE
    \STATE Compute the SWA output $\bm o^{(l,t)}=\bm o^{(l,t)}_\text{SWA}=\mathcal A^{(l)}_\text{SWA}(\bm q_t^{(l)}, \mathcal C^{(l)})$;
    \ENDIF
    \STATE Apply the FFN layer to $\bm o^{(l,t)}$;
    \ENDFOR
    \STATE Generate the next token and append to $R$;
    \ENDWHILE
    \RETURN The detokenized sequence $R$;
    \end{algorithmic}
\end{algorithm}

\section{Details about the Kernel}
\label{app:kernel}
\camrev{The prefill implementation compacts queries routed to the FullAttn and SWA branches into separate, position-sorted lists and tiles them across query positions. This decouples arithmetic intensity from the GQA group size, avoiding a constraint common to many sparse-attention kernels. Figure \ref{fig:swiattn_kernel} provides a compact TileLang implementation.}

\begin{figure*}[!t]
\centering
\begin{lstlisting}[style=appendixcode]
import tilelang
from tilelang import language as T

PASS_CONFIGS = {
    tilelang.PassConfigKey.TL_DISABLE_TMA_LOWER: True,
    tilelang.PassConfigKey.TL_DISABLE_WARP_SPECIALIZED: True,
    tilelang.PassConfigKey.TL_ENABLE_FAST_MATH: True,
}

@tilelang.jit(pass_configs=PASS_CONFIGS)
def switch_gqa_prefill_branch(
    Q, K, V, QueryIndices, QueryCount, Output, Lse,
    heads, dim, num_queries, kv_group, window_size,
    sm_scale=None, is_local=False, block_M=64, block_K=64,
    num_stages=2, threads=128,
):
    assert dim == 128 and heads % kv_group == 0
    assert block_M in (64, 128) and block_K in (64, 128)
    assert num_queries % block_M == 0 and window_size > 0
    sm_scale = ((1.0 / dim) ** 0.5 if sm_scale is None else sm_scale) * 1.4426950408889634
    B, S, S_kv = T.dynamic("batch"), T.dynamic("seq_len"), T.dynamic("seq_len_kv")
    BF16, F32, I32 = T.bfloat16, T.float32, T.int32
    BM, BK, D = block_M, block_K, dim
    NM, heads_per_kv = num_queries // BM, heads // kv_group

    Q: T.Tensor([B, S, heads, D], BF16); K: T.Tensor([B, S_kv, kv_group, D], BF16)
    V: T.Tensor([B, S_kv, kv_group, D], BF16)
    QueryIndices: T.Tensor([B, num_queries], I32); QueryCount: T.Tensor([B], I32)
    Output: T.Tensor([B, S, heads, D], BF16); Lse: T.Tensor([B, S, heads], F32)

    with T.Kernel(NM, B, heads, threads=threads) as (bx, by, bz):
        Qs = T.alloc_shared([BM, D], BF16)
        Ks, Vs = T.alloc_shared([BK, D], BF16), T.alloc_shared([BK, D], BF16)
        Ps, Os = T.alloc_shared([BM, BK], BF16), T.alloc_shared([BM, D], BF16)
        qpos, valid = T.alloc_fragment([BM], I32), T.alloc_fragment([BM], "bool")
        acc_s = T.alloc_fragment([BM, BK], F32); acc_o = T.alloc_fragment([BM, D], F32)
        row_max = T.alloc_fragment([BM], F32); row_max_prev = T.alloc_fragment([BM], F32)
        row_sum = T.alloc_fragment([BM], F32); row_sum_block = T.alloc_fragment([BM], F32)
        alpha = T.alloc_fragment([BM], F32)

        b, h, kv = by, bz, bz // heads_per_kv
        q0, count = bx * BM, QueryCount[by]
        edge = T.max(count - 1, 0)
        first_slot, last_slot = T.min(q0, edge), T.min(q0 + BM - 1, edge)
        first_q = T.max(QueryIndices[b, first_slot], 0)
        last_q = T.max(QueryIndices[b, last_slot], 0)
        first_key = T.max(first_q - window_size + 1, 0) if is_local else 0
        first_block = first_key // BK
        last_block = T.ceildiv(T.min(last_q + 1, S_kv), BK)
        num_blocks = T.max(last_block - first_block, 0)

        for mi in T.Parallel(BM):
            slot = q0 + mi; valid[mi] = slot < count
            qpos[mi] = T.if_then_else(valid[mi], QueryIndices[b, slot], 0)
        for mi in T.serial(BM):
            q = T.max(0, T.min(qpos[mi], S - 1))
            T.copy(Q[b, q, h, :], Qs[mi, :], disable_tma=True)

        T.fill(acc_o, 0); T.fill(row_sum, 0); T.fill(row_max, -(2**30))
        for jb in T.Pipelined(num_blocks, num_stages=num_stages):
            key0 = (first_block + jb) * BK
            T.copy(K[b, key0:key0 + BK, kv, :], Ks, disable_tma=True)
            for mi, ni in T.Parallel(BM, BK):
                key_pos = key0 + ni
                lower = T.max(qpos[mi] - window_size + 1, 0) if is_local else 0
                legal = valid[mi] and key_pos < S_kv and key_pos >= lower and key_pos <= qpos[mi]
                acc_s[mi, ni] = T.if_then_else(legal, 0, -T.infinity(F32))

            T.wgmma_gemm(Qs, Ks, acc_s, transpose_B=True, policy=T.GemmWarpPolicy.FullRow)
            T.wait_wgmma(0); T.copy(row_max, row_max_prev)
            T.reduce_max(acc_s, row_max, dim=1, clear=False)
            for mi in T.Parallel(BM):
                row_max[mi] = T.max(row_max[mi], row_max_prev[mi])
                alpha[mi] = T.exp2((row_max_prev[mi] - row_max[mi]) * sm_scale)
            for mi, ni in T.Parallel(BM, BK):
                acc_s[mi, ni] = T.exp2((acc_s[mi, ni] - row_max[mi]) * sm_scale)
            T.reduce_sum(acc_s, row_sum_block, dim=1)
            for mi in T.Parallel(BM):
                row_sum[mi] = row_sum[mi] * alpha[mi] + row_sum_block[mi]
            for mi, di in T.Parallel(BM, D): acc_o[mi, di] *= alpha[mi]
            T.copy(acc_s, Ps); T.copy(V[b, key0:key0 + BK, kv, :], Vs, disable_tma=True)
            T.wgmma_gemm(Ps, Vs, acc_o, policy=T.GemmWarpPolicy.FullRow); T.wait_wgmma(0)

        for mi, di in T.Parallel(BM, D): acc_o[mi, di] /= row_sum[mi]
        for mi in T.Parallel(BM):
            row_sum[mi] = T.log2(row_sum[mi]) + row_max[mi] * sm_scale
        T.copy(acc_o, Os)
        for mi, di in T.Parallel(BM, D):
            if valid[mi]: Output[b, qpos[mi], h, di] = Os[mi, di]
        for mi in T.Parallel(BM):
            if valid[mi]: Lse[b, qpos[mi], h] = row_sum[mi]
\end{lstlisting}
\vspace{-2mm}
\caption{\camrev{Compact TileLang implementation of the branch-specific \method{} prefill kernel.}}
\label{fig:swiattn_kernel}
\end{figure*}

\begin{camrevblock}
\section{Motivation for Adaptive Regularization}
\label{app:adareg}
The adaptive weight in Eq.\ref{eq:ada-weight} is motivated by two complementary intuitions. First, if the mapping from the $l$-th layer output to the final language-modeling loss is locally Lipschitz continuous \cite{virmaux2018lipschitz}, a larger discrepancy between the FullAttn and SWA outputs may correspond to a larger change in the downstream loss. This motivates using branch disagreement as a signal for weakening the efficiency-oriented penalty:
\[
\begin{aligned}
&|\mathcal{L}_{\text{LM}}(\bm{o}_{\text{SWA}}^{(t,l)})
- \mathcal{L}_{\text{LM}}(\bm{o}_{\text{FULL}}^{(t,l)})| \\
&\qquad\le K \norm{\bm o_{\text{FULL}}^{(t,l)}-\bm o_{\text{SWA}}^{(t,l)}}_2 .
\end{aligned}
\]
Second, the information-bottleneck perspective \cite{tishby2000information, alemi2016deep} views token-level NLL as an indicator of prediction difficulty and information demand. A higher NLL therefore motivates preserving greater flexibility to access global context. Combining NLL with branch disagreement allows the regularization strength to adapt to both token difficulty and the difference between the two attention branches, encouraging SWA for locally predictable tokens while retaining FullAttn as an option when local information may be insufficient.
\end{camrevblock}

\section{Details about the Datasets}
\label{app:dataset}

In this section, we provide comprehensive descriptions of the datasets used in our experiments, organized by the three evaluation categories: commonsense reasoning, in-context retrieval, and long-context understanding.

\subsection{Commonsense Reasoning}
\label{app:commonsense}

This category evaluates models' ability to perform reasoning based on everyday world knowledge.

\begin{itemize}[leftmargin=*]
\setlength\itemsep{0em}
    \item \textbf{Lambada (LMB.)} \citep{paperno2016lambada}: A word prediction task requiring broad discourse context understanding. Models must predict the final word of a passage, which cannot be inferred from local context alone. The dataset contains passages extracted from BookCorpus. We report perplexity on the standard test set.
    
    \item \textbf{Wikitext (Wiki.)} \citep{merity2016pointer}: A language modeling dataset derived from Wikipedia articles. We evaluate perplexity on the test split to assess generative capability over coherent, long-form text.
    
    \item \textbf{BoolQ} \citep{clark2019boolq}: A binary question answering dataset where questions are naturally occurring yes/no queries from Google search. Evaluation is based on exact match accuracy.
    
    \item \textbf{PIQA} \citep{bisk2020piqa}: Physical Interaction QA, focusing on everyday physical reasoning. Models choose between two plausible solutions to a goal. Accuracy is reported on the test set.
    
    \item \textbf{SIQA} \citep{sap2019social}: Social Interaction QA, assessing understanding of social commonsense. Models select the most appropriate response to a given situation among candidates.
    
    \item \textbf{HellaSwag (Hella.)} \citep{zellers2019hellaswag}: A challenging dataset for commonsense natural language inference (NLI), requiring models to predict the most plausible continuation of an event description among four candidates. Normalized accuracy is evaluated on the test split.
    
    \item \textbf{WinoGrande (Wino.)} \citep{sakaguchi2021winogrande}: A 44K-example Winograd schema dataset testing pronoun disambiguation with adversarial filtering. We report binary accuracy.
    
    \item \textbf{ARC-easy \& ARC-challenge (ARC-e/c)} \citep{clark2018think}: The AI2 Reasoning Challenge, containing grade-school science questions. ARC-e consists of relatively straightforward questions, while ARC-c includes more challenging ones requiring multi-step reasoning. Accuracy is reported for ARC-e, while normalized accuracy is reported for ARC-c.
    
    \item \textbf{OpenBookQA (OBQA)} \citep{mihaylov2018can}: A dataset requiring external commonsense knowledge beyond the provided ``open book'' fact. Models answer multiple-choice science questions; evaluation is based on normalized accuracy.
\end{itemize}

\subsection{In-Context Retrieval}
\label{app:retrieval}

Following \citet{arora2024just}, this category assesses models' ability to retrieve and utilize relevant information from a maximum of $4,096$ token context. We use Cloze Completion Formatting prompts aligned with next-word prediction.

\begin{itemize}[leftmargin=*]
\setlength\itemsep{0em}
    \item \textbf{SWDE} \citep{lockard2019openceres}: A structured extraction dataset where models answer questions by retrieving attribute values from semi-structured web pages (HTML).
    
    \item \textbf{FDA} \citep{arora2023language}: Food and Drug Administration document QA, requiring extraction of specific regulatory information from PDF documents.
    
    \item \textbf{TriviaQA (TQA)} \citep{joshi2017triviaqa}: A large-scale reading comprehension dataset with question-answer-evidence triples.
    
    \item \textbf{Natural Questions (NQ)} \citep{kwiatkowski2019natural}: Real user queries to Google Search with answers annotated on Wikipedia pages.
    
    \item \textbf{DROP} \citep{dua2019drop}: A dataset requiring discrete reasoning over numerical and temporal information. Models generate answers via span extraction, counting, or arithmetic.
    
    \item \textbf{SQuAD (SQD)} \citep{rajpurkar2018know}: Stanford Question Answering Dataset, where answers are text spans from Wikipedia passages. 
\end{itemize}

\subsection{Long-Context Understanding}
\label{app:longbench}

\camrev{We adopt benchmark datasets from LongBench-E \citep{bai2024longbench} to evaluate performance on tasks requiring processing of extended contexts. We focus on single-document QA, summarization, and few-shot learning, which cover natural-language document understanding and in-context learning. Specifically, we include MultiFieldQA-en, Qasper, GovReport, MultiNews, TriviaQA, SAMSum, and TREC. We exclude the multi-document QA datasets HotpotQA and 2WikiMultihopQA, the synthetic datasets PassageCount and PassageRetrieval-en, and the code-completion datasets LCC and RepoBench-P. The included datasets are described below.}

\begin{itemize}[leftmargin=*]
\setlength\itemsep{0em}
    \item \textbf{MultiFieldQA (MFQA)} \citep{bai2024longbench}: MFQA is a multi-domain QA dataset with long documents spanning legal, scientific, judicial, and government fields. Answers require synthesizing information across distant passages.
    
    \item \textbf{TRec} \citep{li2002learning}: TRec is a question classification task requiring models to categorize questions into 50 predefined topic classes.
    
    \item \textbf{MultiNews (MNs)} \citep{fabbri2019multi}: MultiNews is a multi-document summarization dataset containing news articles about the same event.
    
    \item \textbf{GovReport (GvR)} \citep{huang2021efficient}: GovReport is a long-document summarization task based on U.S. government reports and congressional hearings.
    
    \item \textbf{Qasper (Qas)} \citep{dasigi2021dataset}: Qasper is a QA dataset over NLP research papers, featuring questions requiring understanding of methodology, results, and claims.
    
    \item \textbf{TriviaQA (TQA)} \citep{joshi2017triviaqa}: The TriviaQA dataset in LongBench-E is a long-context variant of the original version, where relevant evidence is embedded within extended passages to test retrieval under length pressure.

    \item \textbf{SAMSum (SSM)} \citep{gliwa2019samsum}: The SAMSum dataset consists of messenger-like conversations as the input with summaries annotated by humans as the target.
\end{itemize}

\subsection{Evaluation Metrics}
We adopt different evaluation metrics for these datasets. Specifically, for Lambada and Wikitext, we use perplexity. For BoolQ, PIQA, SIQA, WinoGrande, and ARC-e, we use accuracy. For HellaSwag, ARC-c, and OpenBookQA, we use normalized accuracy. For in-context retrieval tasks, we use the ``contains" metric and follow the prompt template of \citealt{arora2024just}. For LongBench-E, we use the default evaluation metric provided in \citealt{bai2024longbench} with implementation from \citealt{eval-harness}.

\section{Details about the Baselines}
\label{app:baseline}
In this part, we provide more details about the implementation of the baseline models.
\begin{itemize}[leftmargin=*]
\setlength\itemsep{0em}
\item \textbf{SWA-ZS.} The zero-shot sliding window attention baseline is implemented using the corresponding full attention model. It directly restricts the attention span of full attention transformers to $1024$ in alignment with \method{}.
\item \textbf{SWA-CPT.} The continually pretrained sliding window attention model is obtained by performing CPT on the SWA-ZS model. We use the same amount of tokens as the proposed \method{} to train this baseline model.
\item \camrev{\textbf{SH (3:1).} This StaticHybrid baseline alternates one full attention layer after every three SWA layers. In the $22$-layer transformer model, the $4,8,12,16,20$-th layers use full attention, while the remaining layers use SWA.}
\item \camrev{\textbf{SH (6:1).} This StaticHybrid baseline uses three full attention layers and $19$ SWA layers, giving an approximate SWA-to-FullAttn layer ratio of $6{:}1$. Specifically, the $7,14,21$-st layers use full attention.}
\item \camrev{\textbf{DuoAttention.} We evaluate DuoAttention \citep{xiao2025duoattention} using the same $1.5$B backbone and LongBench-E evaluation setting.}
\item \textbf{FullAttn.} The full attention baseline is a $1.5$B language model with $22$ transformer layers. The hidden size is set to $2048$. The standard version is trained on $240$B tokens of $4$K context lengths, while the $32$K version is trained on $192$B tokens of $4$K lengths and $24$B tokens of $32$K lengths.
\end{itemize}

\begin{figure*}[ht]
    \centering
    \includegraphics[width=\linewidth]{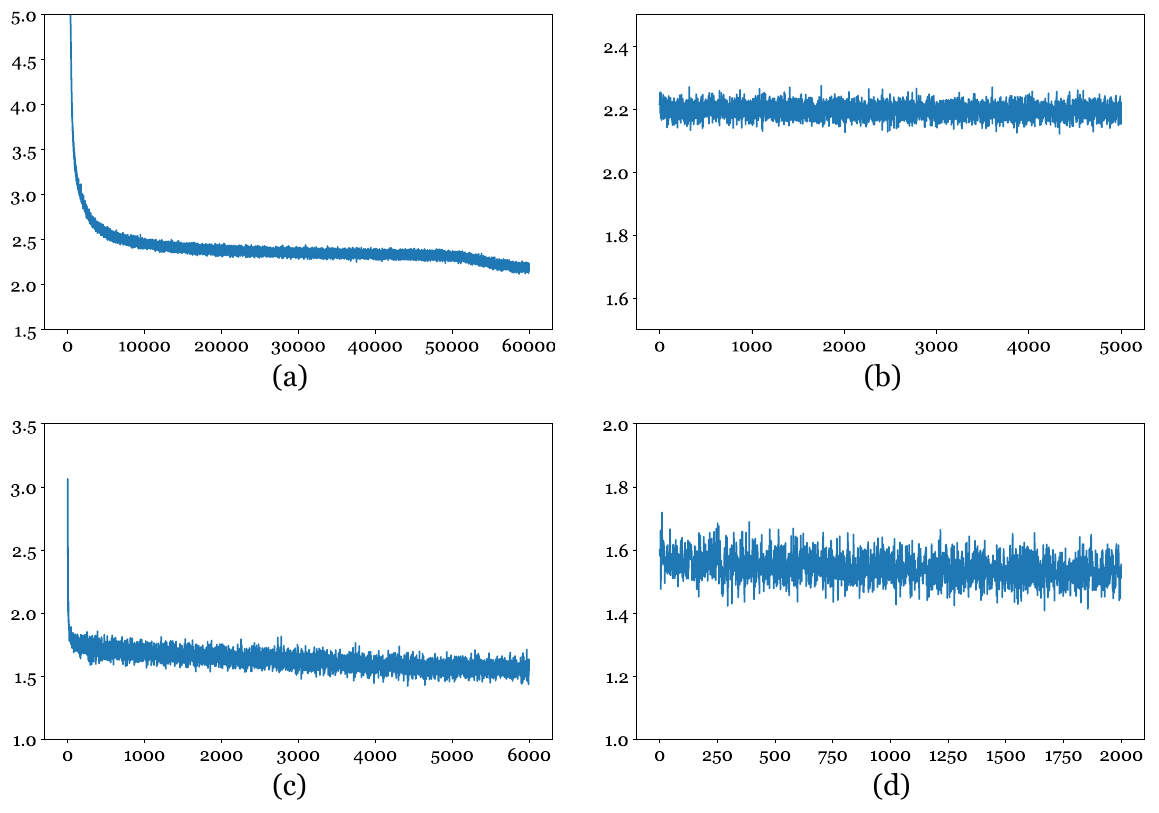}
    \caption{Language modeling loss curves of (a) pretraining of the full attention transformer on $4$K-length sequences, (b) continual pretraining of \method{} on $4$K-length sequences, (c) pretraining of the full attention transformer on $32$K-length sequences, and (d) continual pretraining of \method{} on $32$K-length sequences.}
    \label{fig:loss-curves}
\end{figure*}

\section{Additional Implementation Details}
\label{app:implementation}
In the experiments, we adopt a $1.5$B transformer model with $22$ layers and the hidden dimension of $2048$. For the attention module, group query attention \cite{ainslie2023gqa} is used with $16$ attention heads and $8$ key-value heads. The dimension of each query, key, and value vectors is set to $128$. In dual-branch routing, the router function $\mathcal R(\cdot)$ is applied after the root-mean-square (RMS) layer normalization \cite{zhang2019root}. 

\camrev{Within each context-length setting, FullAttn pretraining and \method{} CPT use the same data distribution. The $4$K corpus mainly contains English and Chinese text collected from web pages, books, encyclopedias, and other general sources, together with some multilingual and domain-specific data. The $32$K corpus contains more long-form and instruction data; about $20\%$ is instruction data, including diverse questions and answers with varying reasoning lengths.}

For the $4$K-length model, we first pretrain the full attention transformers with $240$B tokens, where we set the batch size to $1024$ and train for $60000$ steps. During the first $2000$ steps of the pretraining, we linearly warm up the learning rate from $0$ to $1\times 10^{-3}$. During the last $12000$ steps, we decay the learning rate to $0$ with a cosine schedule. For the continual pretraining of \method{}, we first initialize the model as described in Section \ref{sec:cpt} and then continually pretrain the model for another $20$B tokens with the same batch size for $5000$ steps. During the CPT stage, we linearly warm up the learning rate from $0$ to $1\times 10^{-4}$ for $500$ steps and decay the learning rate in the remaining $4500$ steps with a cosine schedule.

\begin{figure*}
    \centering
    \includegraphics[width=\linewidth]{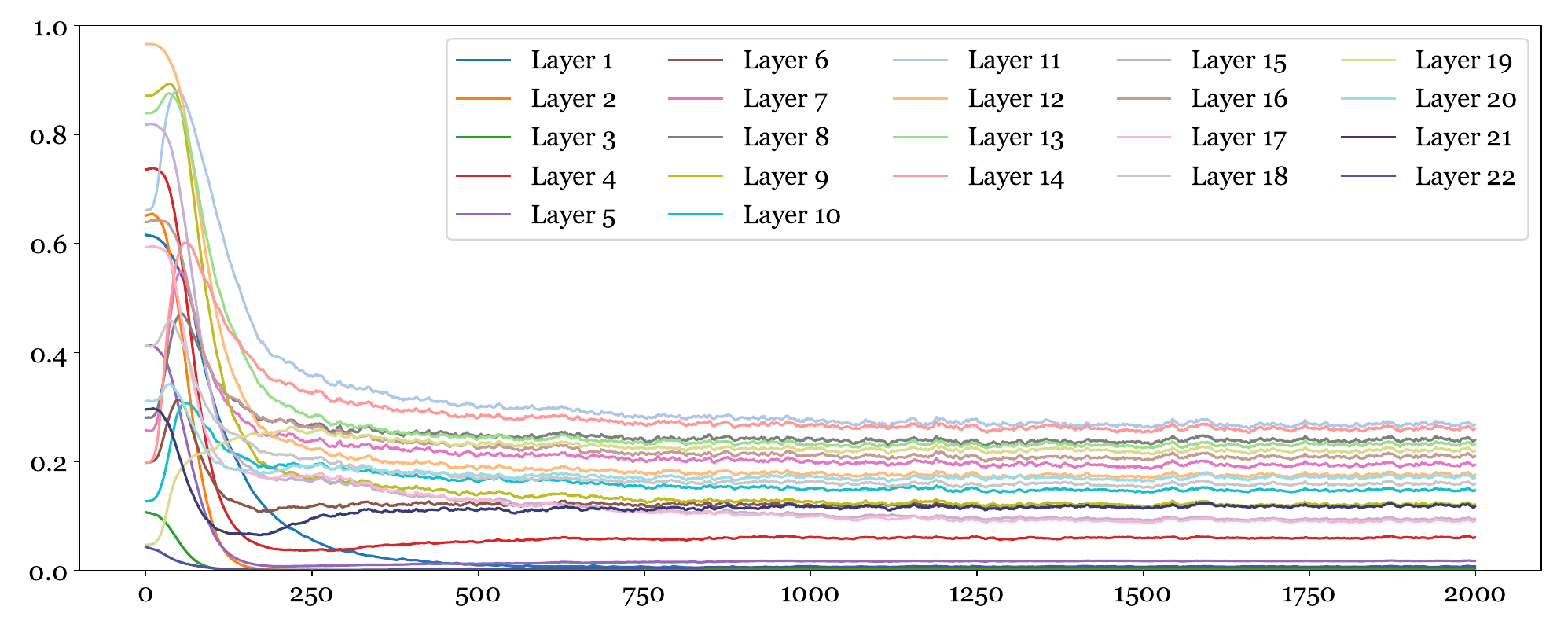}
    \caption{The changes of full attention ratio during the continual pretraining stage. The ratio is averaged within each transformer layer.}
    \label{fig:fr-all}
\end{figure*}

For the $32$K-length version of the model, we first pretrain the full attention transformer model for $192$B tokens on $4$K-length text, and then extend the context length to $32$K using YaRN \cite{peng2023yarn}. Then we train the full attention model on $32$K-length texts for another $24$B tokens with the batch size of $128$ for $6000$ steps. During the first $2000$ steps, the learning rate linearly warms up to $1\times 10^{-3}$, and during the last $24$B token on the $32$K-length texts (\emph{i.e.}, last $6000$ steps), the learning rate is decayed to $0$ under a cosine schedule. The $32$K-length version of \method{} is first initialized using the aforementioned full attention model as described in Section \ref{sec:cpt} and then continually pretrained for another $8$B tokens with the same batch size (\emph{i.e.}, $128$) for another $2000$ steps. In the first $200$ steps, we warm up the learning rate from $0$ to $2\times 10^{-4}$, and in the remaining $1800$ steps, we decay the learning rate to $0$ using a cosine schedule. \camrev{The additional CPT budgets correspond to $8.3\%$ of the preceding $4$K pretraining budget and $3.7\%$ of the preceding $32$K model-training budget. CPT is used to adapt the architecture and learn the newly introduced router, and SWA-CPT and the StaticHybrid baselines receive the same additional token budget as \method{}.}

We use Megatron-LM \cite{shoeybi2019megatron} to train the transformer model and adopt LM-Eval-Harness \cite{eval-harness} to evaluate the models on various datasets. By default, the results are the models' performance under a single run. For the NIAH experiments, the RULER \cite{hsieh2024ruler} setup is applied. All experiments are performed on Ascend 910C NPUs.

\section{Loss Curves}
In this part, we provide the loss curves of model optimization. Specifically, we visualize the language modeling loss of (a) pretraining of the full attention transformer on $4$K-length sequences; (b) continual pretraining of \method{} on $4$K-length sequences; (c) pretraining of the full attention transformer on $32$K-length sequences; (d) continual pretraining of \method{} on $32$K-length sequences. The results are presented in Figure \ref{fig:loss-curves}. These curves suggest that the optimization procedure of \method{} is relatively stable with no loss spikes. Furthermore, the transition during the continual pretraining stage of \method{} demonstrates convergence dynamics that are highly competitive with the standard baseline. This consistent stability across both 4K and 32K context windows confirms that CPT serves as a highly reliable and efficient approach for seamlessly converting standard full attention models into the proposed \method{} architecture, without sacrificing optimization integrity.

\begin{figure}
    \centering
    \includegraphics[width=\linewidth]{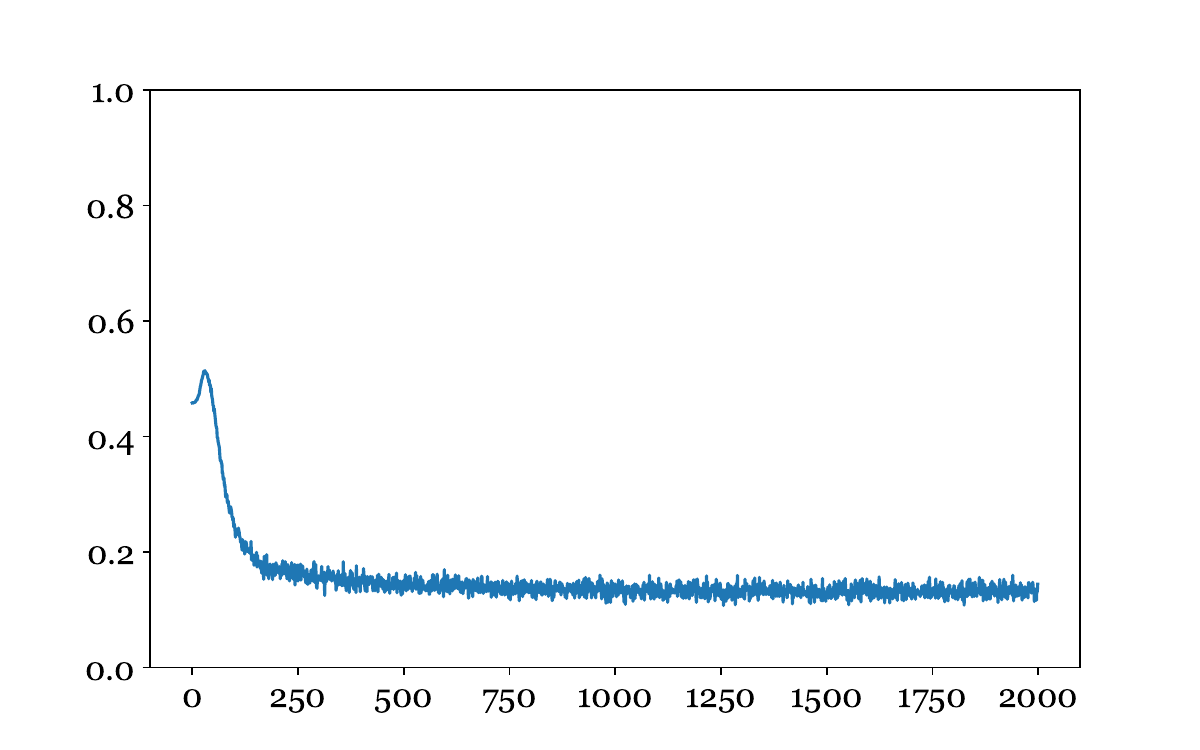}
    \caption{The changes of full attention ratio during the continual pretraining stage. The averaged ratio over all transformer layers is reported.}
    \label{fig:fr-avg}
\end{figure}

\section{Changes of Full Attention Ratio during Continual Pretraining}
In this part, we present the changes of the full attention ratios during the continual pretraining stage. Specifically, we measure the frequency of the router choosing the full attention branch during CPT on the pretraining data. The averaged ratio of full attention is presented in Figure \ref{fig:fr-avg}, while the averaged full attention ratio of each layer is visualized in Figure \ref{fig:fr-all} (where the curves are smoothed for better visualization). From the results, we can see that the proposed \method{} chooses the full attention branch with a frequency of around $0.13$, which is more efficient compared to common static hybrid methods (\emph{i.e.}, around $0.25$ full attention ratio). Moreover, the per-layer statistics of \method{} suggest that different layers exhibit distinct behaviors: some layers (\emph{e.g.}, Layer 8, 11, 13, 14) tend to use full attention more frequently, whereas others (\emph{e.g.}, Layer 1, 2, 3, 22) tend to use SWA more frequently. Moreover, these results show that the changes of full attention ratio are relatively smooth, and that the ratio stabilizes after around $500$ iterations.

\begin{camrevblock}
\begin{table*}[t]
\centering
\caption{\rev{Performance of 3B models on common sense reasoning datasets.}}
\label{tab:revcs}
\resizebox{\textwidth}{!}{%
\def\arraystretch{1.2}
\renewcommand\tabcolsep{5pt}
\begin{tabular}{
  >{\raggedright\arraybackslash}p{2cm}
  !{\vrule width 1.0pt}
  *{2}{>{\centering\arraybackslash}p{1.1cm}}
  >{\centering\arraybackslash}p{1.1cm}
  !{\vrule width 1.0pt}
  *{8}{>{\centering\arraybackslash}p{1.1cm}}
  >{\centering\arraybackslash}p{1.1cm}
}
\Xhline{1.3pt}
\rowcolor{tableheadcolor!50} Datasets & LMB. & Wiki. & AVG & BoolQ & PIQA & SIQA & Hella. & Wino. & ARC-e & ARC-c & OBQA & AVG \\
\Xhline{1.0pt}
\rowcolor{tableheadcolor!50} Metrics
& ppl$\downarrow$ & ppl$\downarrow$ & ppl$\downarrow$ & acc$\uparrow$ & acc$\uparrow$ & acc$\uparrow$ & acc\_n$\uparrow$ & acc$\uparrow$ & acc$\uparrow$ & acc\_n$\uparrow$ & acc\_n$\uparrow$ & acc$\uparrow$ \\
\Xhline{1.3pt}
FullAttn & 6.14 & 13.8 & 9.99 & 63.3 & 74.8 & 43.9 & 63.5 & 61.1 & 66.3 & 35.4 & 39.0 & 55.9 \\
\Xhline{1.0pt}
\method{} & 6.19 & 13.8 & 9.99 & 65.6 & 74.8 & 43.4 & 63.5 & 61.7 & 65.0 & 34.6 & 38.4 & 55.9 \\
\Xhline{1.3pt}
\end{tabular}}
\end{table*}

\begin{table*}[t]
\centering
\caption{\rev{Performance of 3B models on in-context retrieval and LongBench-E datasets.}}
\label{tab:revlb}
\resizebox{\textwidth}{!}{%
\def\arraystretch{1.2}
\renewcommand\tabcolsep{5pt}
\begin{tabular}{
  >{\raggedright\arraybackslash}p{2cm}
  !{\vrule width 1.0pt}
  *{7}{>{\centering\arraybackslash}p{1.1cm}}
  !{\vrule width 1.0pt}
  *{6}{>{\centering\arraybackslash}p{1.1cm}}
}
\Xhline{1.3pt}
\rowcolor{tableheadcolor!50} & \multicolumn{7}{c!{\vrule width 1.0pt}}{In-Context Retrieval} & \multicolumn{6}{c}{LongBench-E} \\
\Xhline{1.0pt}
\rowcolor{tableheadcolor!50} Methods & SWDE & FDA & TQA & NQ & DROP & SQD & AVG & GvR & MNs & MFQA & Qas & TQA & AVG \\
\Xhline{1.3pt}
FullAttn & 65.4 & 76.5 & 65.0 & 38.0 & 34.4 & 53.3 & 55.4 & 27.3 & 21.5 & 39.6 & 27.6 & 70.0 & 37.2 \\
\Xhline{1.0pt}
\method{} & 66.3 & 77.7 & 65.6 & 38.1 & 33.5 & 53.7 & 55.8 & 26.8 & 20.2 & 39.7 & 26.6 & 69.5 & 36.6 \\
\Xhline{1.3pt}
\end{tabular}}
\end{table*}

\section{Experiments on 3B Models}
\label{app:3b}
We also evaluate a 3B transformer with $28$ layers, hidden size $2048$, $16$ attention heads, $4$ key-value heads, and FFN size $9216$. We reuse the 1.5B training schedule and hyperparameters without a scale-specific search. Tables \ref{tab:revcs} and \ref{tab:revlb} report commonsense reasoning, in-context retrieval, and long-context understanding results for FullAttn and \method{}, providing preliminary evidence at a larger scale.
\end{camrevblock}

\section{Additional NIAH Results}
We compare aggregate NIAH accuracy under the same evaluation setting in Table \ref{app:tab:niah}. FullAttn and \method{} both achieve perfect retrieval accuracy, while SWA and SH (3:1) obtain lower scores.

\begin{table}[H]
\centering
\caption{Aggregate NIAH retrieval accuracy under the same evaluation setting.}
\label{app:tab:niah}
\vspace{-2mm}
\footnotesize
\begin{tabular}{lc}
\Xhline{1.3pt}
\rowcolor{tableheadcolor!50} Method & NIAH Accuracy \\
\Xhline{1.3pt}
FullAttn & 1.00 \\
\rowcolor{tablerowcolor}\method{} & 1.00 \\
SWA & 0.20 \\
\rowcolor{tablerowcolor}SH (3:1) & 0.95 \\
\Xhline{1.3pt}
\end{tabular}
\vspace{-2mm}
\end{table}

\section{Additional Results on Efficiency}
\label{app:efficiency}

\begin{camrevblock}
\subsection{Training Overheads}
\begin{table}[H]
\centering
\caption{Comparison of runtime per layer during continual pretraining (CPT) with a sequence length of 32K.}
\label{tab:cpt_overhead}
\resizebox{\linewidth}{!}{
\def\arraystretch{1.2}
\renewcommand\tabcolsep{5pt}
\begin{tabular}{
  >{\raggedright\arraybackslash}p{2.5cm}
  >{\centering\arraybackslash}p{4cm}
  >{\centering\arraybackslash}p{4cm}
}
\Xhline{1.3pt}
\rowcolor{tableheadcolor!50} Methods & Forward time (s) & Backward time (s) \\
\Xhline{1.3pt}
FullAttn & 0.0595 & 0.1459 \\
\Xhline{1.0pt}
\method{} & 0.0645 & 0.1574 \\
\Xhline{1.0pt}
Overhead & 8.4\% & 7.9\% \\
\Xhline{1.3pt}
\end{tabular}}
\end{table}
In this part, we analyze the computational overhead introduced during CPT. The SWA branch is computed efficiently, while the two attention branches share the QKV projections and the remaining model components, including the FFN layers. Consequently, computing both attention branches does not double the per-layer runtime.

Table~\ref{tab:cpt_overhead} reports the runtime per layer during CPT at a sequence length of 32K. Compared with FullAttn, \method{} introduces $8.4\%$ forward and $7.9\%$ backward overhead.
\end{camrevblock}

\subsection{\camrev{Efficiency Comparison with SH (3:1)}}
\begin{table}[H]
\centering
\caption{\camrev{Efficiency comparison between \method{} and the SH (3:1) baseline.}}
\label{tab:statichybrid_efficiency}
\resizebox{\linewidth}{!}{%
\def\arraystretch{1.2}
\renewcommand\tabcolsep{4pt} 
\begin{tabular}{
  >{\raggedright\arraybackslash}p{1.8cm}
  >{\centering\arraybackslash}p{1.4cm}
  >{\centering\arraybackslash}p{1.4cm}
  >{\centering\arraybackslash}p{1.4cm}
  >{\centering\arraybackslash}p{1.4cm}
  >{\centering\arraybackslash}p{2.2cm}
}
\Xhline{1.3pt}
\rowcolor{tableheadcolor!50} & \multicolumn{2}{c}{Standard Setting} & \multicolumn{2}{c}{Simple Setting} & \\
\Xcline{2-5}{1.0pt} 
\rowcolor{tableheadcolor!50} \multirow{-2}{*}{Methods} & Memory & GFLOPs & Memory & GFLOPs & \multirow{-2}{*}{LongBench-E} \\
\Xhline{1.3pt}
\camrev{SH (3:1)} & 8.0K & 1.38 & 8.0K & 1.38 & 35.2 \\
\Xhline{1.0pt}
\method{} & \textbf{7.3K} & \textbf{1.26} & \textbf{2.6K} & \textbf{0.44} & \textbf{37.4} \\
\Xhline{1.3pt}
\end{tabular}}
\end{table}

\camrev{We further provide an efficiency comparison between \method{} and the SH (3:1) baseline. Table~\ref{tab:statichybrid_efficiency} reports the memory access (in terms of tokens) and GFLOPs under both the standard and simple settings (as described in Section 4.4 of the main paper), alongside their average performance on the LongBench-E benchmark. }

\camrev{The results show that \method{} is more efficient than SH (3:1), especially in the simple setting: it reduces memory access and GFLOPs by over 67\% and improves LongBench-E performance from 35.2 to 37.4.}

\rev{Additionally, \method{} is complementary to existing efficiency techniques: it can replace StaticHybrid's full-attention layers or equip a sparse-attention mechanism with a router that decides whether to scan global memory.}

\begin{camrevblock}
\section{Statistical Significance}
\begin{table}[H]
\centering
\caption{Performance comparison over three runs and statistical significance ($p$-value) across commonsense reasoning, in-context retrieval, and LongBench-E.}
\label{tab:performance_significance}
\vspace{-2mm}
\resizebox{\linewidth}{!}{%
\def\arraystretch{1.2}
\renewcommand\tabcolsep{5pt}
\begin{tabular}{
  >{\raggedright\arraybackslash}p{2cm}
  >{\centering\arraybackslash}p{2.5cm}
  >{\centering\arraybackslash}p{2.5cm}
  >{\centering\arraybackslash}p{2.5cm}
}
\Xhline{1.3pt}
\rowcolor{tableheadcolor!50} Methods & Commonsense & Retrieval & LongBench \\
\Xhline{1.3pt}
SH (3:1) & 53.6$\pm$0.32 & 50.6$\pm$0.09 & 35.2$\pm$0.44 \\
\Xhline{1.0pt}
FullAttn & -- & -- & 37.4$\pm$0.18 \\
\Xhline{1.0pt}
\method{} & 53.7$\pm$0.30 & 52.9$\pm$0.11 & 37.4$\pm$0.35 \\
\Xhline{1.0pt}
$p$-value (vs. SH) & $0.713$ & $<0.001$ & $<0.01$ \\
\Xhline{1.0pt}
$p$-value (vs. FullAttn) & -- & -- & $0.941$ \\
\Xhline{1.3pt}
\end{tabular}}
\vspace{-1mm}
\end{table}
Table \ref{tab:performance_significance} reports results over three runs and the corresponding two-sided Welch's $t$-tests. The improvements of \method{} over SH (3:1) are statistically significant on in-context retrieval and LongBench-E. On LongBench-E, the comparison with FullAttn gives $p=0.941$, supporting performance parity.
\end{camrevblock}

\end{document}